\documentclass[lettersize,journal]{IEEEtran}
\usepackage{amsmath,amsfonts}
\usepackage{algorithmic}
\usepackage{algorithm}
\usepackage{amsmath}
\usepackage{mathtools}
\usepackage{array}
\usepackage{amssymb}
\usepackage[none]{hyphenat}
\usepackage{hyperref}
\usepackage{cleveref}
\usepackage[caption=false,font=normalsize,labelfont=sf,textfont=sf]{subfig}
\usepackage{hyperref}
\usepackage{textcomp}
\usepackage{stfloats}
\usepackage{url}
\usepackage{verbatim}
\usepackage{graphicx}
\usepackage{pifont}
\usepackage{cite}
\crefname{figure}{Fig.}{Figs.}
\crefname{table}{Table}{Tables}
\crefname{section}{Section}{Sections}
\crefname{algorithm}{Algorithm}{Algorithms}

\newcommand{\xmarkraw} {\textbf{\ding{55}}}
\newcommand{\cmarkraw} {\textbf{\checkmark}}
\newcommand{\xmark} {\multicolumn{1}{c}{\textbf{\ding{55}}}}
\newcommand{\cmark} {\multicolumn{1}{c}{\textbf{\checkmark}}}
\newcommand{\none} {\multicolumn{1}{c}{-}}
\newcommand{\noneleft} {\multicolumn{1}{l}{-}}

\newcommand{\totseg}{TotalSegmentator}
\newcommand{\suprem}{U-Net SuPreM}

\newcommand{\sammed}{SAM-Med3D}
\newcommand{\nninter}{nnInteractive}
\newcommand{\vista}{VISTA3D}

\newcommand{\parm}{PARMESAN}
\newcommand{\meow}{MEOW}

\newcommand{\totsegdata}{TotSeg Organ}

\newcommand{\msd}{MSD}

\newcommand{\kits}{KiTS23}

\newcommand{\mousemicro}{Mouse $\mu$CT}
\newcommand{\antbrain}{InsectAnatomy}
\newcommand{\brats}{MSD BRATS}

\begin{document}
\bstctlcite{IEEEexample:BSTcontrol}

\title{\meow: Medical Open-World Interactive Segmentation}
\author{Philip Matthias Winter\textsuperscript{1*}, 
Maria Wimmer\textsuperscript{1*}, 
Bernhard Sch\"{a}fl\textsuperscript{1}, 
David Major\textsuperscript{1}, 
Dimitrios Lenis\textsuperscript{1}, 
Astrid Berg\textsuperscript{1}, 
Theresa Neubauer\textsuperscript{1}, 
Ariharasudhan Muthusami\textsuperscript{1}, 
Johannes Novotny\textsuperscript{1}, 
Georg Zotti\textsuperscript{1}, 
Katja B\"{u}hler\textsuperscript{1}

\thanks{\textsuperscript{1}These authors are with the VRVis GmbH - Vienna Research Center for Visual Computing, Vienna, Austria (e-mail: office@vrvis.at).}
\thanks{\textsuperscript{*}These authors contributed equally to this work.}}

\maketitle

\begin{abstract}
In open-world recognition, machine learning models not only make predictions on unknown structures ("generalized open-set"), but also continually learn and improve from those predictions over time. Interactive segmentation, despite its access to user feedback, has so far remained confined to the open-set setting. In this work we introduce \meow, the first framework that turns non-interactive segmentation backbones into interactive open-world learners without updating learnable model parameters. \meow\ combines parameter-free transductive reasoning with memory consolidation to accumulate knowledge and thereby improve future performance. Focusing on medical data, we show that \meow\ consistently improves its own predictions on classes never seen during training, and transfers zero-shot to unfamiliar domains, including micro-CT data and other imaging modalities. 
\end{abstract}


\section{Introduction}
\label{sec:introduction}

Recent advances in lifelong learning have fundamentally shifted how deep learning models are deployed in dynamic environments where data distributions and task definitions continuously evolve \cite{parisi19survey,wang24survey}. When a static model trained on a closed-world distribution is deployed in the wild, it inevitably encounters out-of-distribution data. Processing these unseen distributions without adaptation degrades predictive performance \cite{liu26oodsafety}. In critical vision-tasks \cite{zhao24vision}, e.g. medical image analysis \cite{gutbrod25openmibood} and autonomous driving \cite{shoeb25carOOD}, this may lead to fatal outcomes. Consequently, enabling models to continuously adapt during their operational lifetime is crucial for maintaining high reliability and performance.

However, model adaptation introduces various challenges concerning plasticity-stability trade-offs and catastrophic forgetting \cite{delange19forget}, as well as model complexity and prohibitive computational costs \cite{rusu22progressivenns}. Another conceptual problem arises from the nature of the incoming data stream. Standard lifelong learning frameworks often assume a well-defined stream where novel tasks or classes are explicitly delineated and labeled by human experts. In practical settings, however, models must be able to handle completely novel, unknown classes and domains that emerge dynamically.

To succeed in such a dynamic environment, open-world (OW) learning has emerged as a highly relevant and pressing challenge. In OW-learning, a model must be able to operate in unknown domains, make predictions for unknown classes, and also actively learn and improve over time. This is usually achieved by leveraging knowledge directly from its own predictions on these unknowns.

In clinical environments, closed-world systems with fixed label spaces fail due to rare pathologies and domain shifts. They often yield hazardous, overconfident predictions when unknown structures are forced into predefined categories \cite{karimi23}. OW-learning addresses this by incorporating these unknowns into a model's taxonomy. Moreover, it mitigates severe data scarcity caused by costly expert annotations, licensing constraints, and strict privacy regulations, as exhaustive labeling of every possible class is no longer required. By dynamically adapting to anatomical variations and unseen distributions, OW frameworks help bridge the gap toward safe, robust clinical deployment.

\begin{figure}[t]
	\centering
	\includegraphics[width=1.\linewidth]{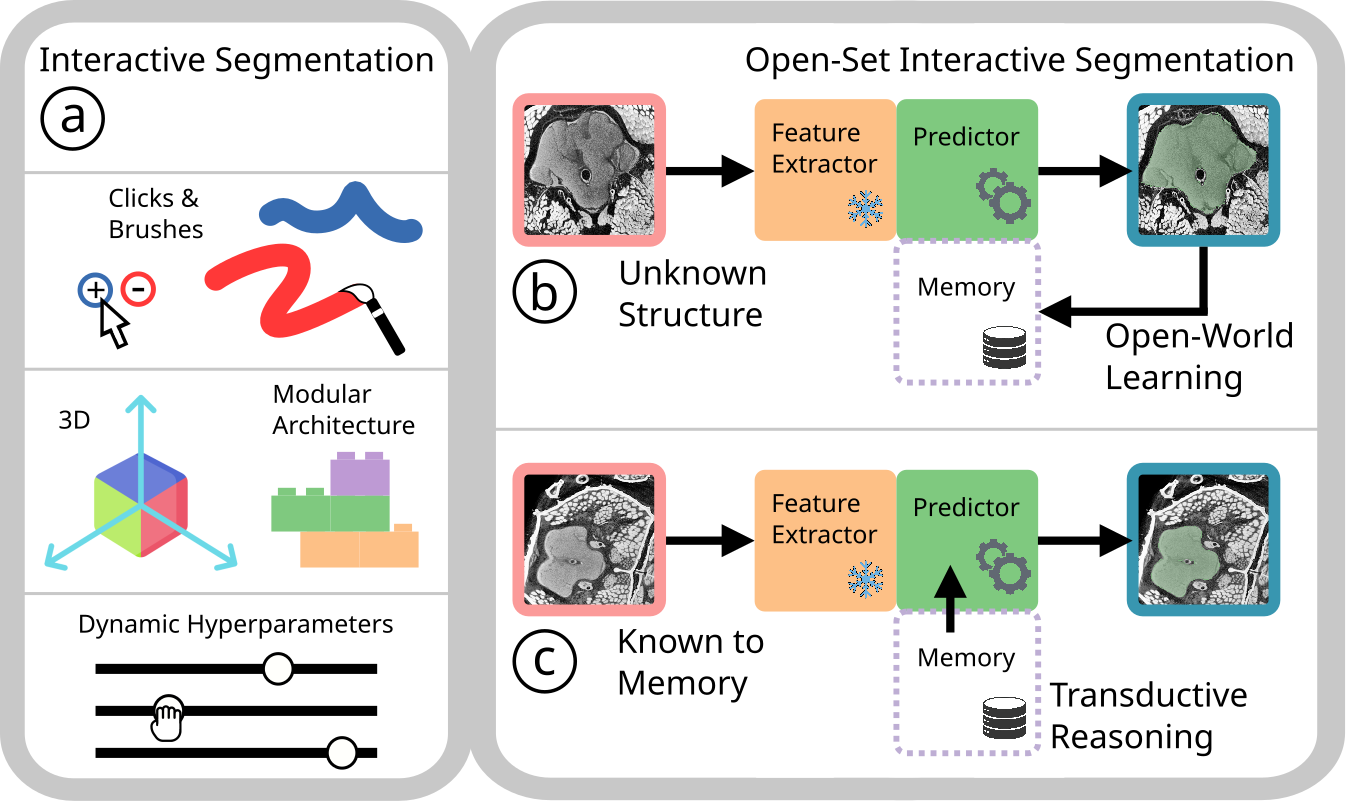}
	\caption{We propose \meow, a framework for open-world interactive segmentation. a) Flexible properties for Interactive Segmentation. b) Segmenting an unknown structure and learning from own prediction. c) Improving segmentation performance by leveraging previously learned structures.}
	\label{fig:eye_catcher}
\end{figure}

OW-learning combines multiple learning paradigms \cite{bendale15openworld}, including incremental, continual, transfer, out-of-distribution, few-shot, and open-set (OS) learning. Furthermore, desired properties encompass mechanisms for fast learning and unlearning at scale. Beyond these, OW systems must unify instance classification for existing classes with novelty detection and label management for new classes. While OS systems are limited to identifying or rejecting unknown classes, OW systems incrementally learn from their own predictions during deployment. Previously unknown classes become known once processed, thereby enhancing future performance. Beyond traditional tasks like classification or regression, OW recognition also extends to generative modeling, where models must adapt to generate samples from previously unseen distributions.

We focus on the 3D vision-task of open-world interactive segmentation (OW-IS), which has not been studied yet. In interactive segmentation (IS) \cite{boykov01graphcut,xu2016deep}, a machine learning (ML) model takes user inputs like clicks or brushes to segment parts of an image. Self-training and pseudo-labeling schemes for OW systems often rely on gradient-based parameter updates, which are prone to confirming a model's systematic biases, degrade representation stability, and require significant computational overhead. This motivates the need for an adaptation mechanism that can exploit a model's own predictions to learn new classes and domains reliably, rapidly, and safely.

In this work, we present the first system capable of OW-IS, thereby addressing the core challenges encountered by an OW-learning system (see \cref{fig:eye_catcher}). On top of the desired OW-properties mentioned above, our approach offers various favorable properties in terms of IS- and OW aspects, including flexible user interactions and a modular architecture. Most importantly, our method requires no gradient-based learning \cite{rumelhart86backprop} to transform non-interactive, frozen-parameter models into flexible IS models while simultaneously enabling efficient OW-learning capabilities. This is achieved by building upon our previously proposed method \parm\ \cite{winwim24}, which is a memory-based approach for solving dense prediction tasks such as semantic segmentation and depth estimation. It was specifically designed as a continual learning method, exhibiting stable knowledge retention without catastrophic forgetting.

Our main contributions are as follows:
\newline
\textbf{Interactive Segmentation:} We introduce \meow, a modular 3D IS system that is capable of handling sparse annotations (e.g., from clicks or brushes) in a flexible manner. In contrast to methods that are trained end-to-end using a fixed interaction strategy during training, our approach takes non-interactive off-the-shelf segmentation models and integrates them into our IS framework. Our system is invariant to diverse annotation styles, ensuring robust performance across different users. Apart from clicking and brushing interactions, our method further supports interactive hyperparameter tuning via sliders.
\newline
\textbf{Open-World Learning:} We propose a solution to OW-IS, i.e., an IS system learning from its own predictions of novel structures. Notably, this also includes the selective removal of previously learned samples and domains. To the best of our knowledge, this is the first operational OW-IS framework.
\newline
\textbf{General Improvements:} To enable efficient OW-IS for 3D data, we introduce modifications to \parm, including improved feature compression, better memory search kernels, and more efficient test-time augmentation. We also improve its message passing approach for post-processing of raw predictions by implementing hierarchical message passing, thereby enhancing overall segmentation quality and robustness. Finally, we propose novel evaluation metrics that are useful for IS.
\newline

Our approach uses publicly available, pre-trained models without requiring any additional training of learnable parameters, making it flexible, modular, updatable, and fast-learning by design. Our work can be fully reproduced with consumer-grade hardware, thereby underlining our energy-efficient use of computational resources.

\section{Related Work}

\subsection{Interactive Segmentation}
Semantic segmentation has recently been transformed by foundation models like \suprem\ \cite{li24unetsuprem}, \totseg\ \cite{wasserthal23totseg}, and MemSAM \cite{deng24memsam}. Being trained on large datasets containing highly diverse classes, these models excel at segmenting basic structures like organs and bones. Interactive segmentation (IS) \cite{marinov24_ISsurvey} is typically addressed via end-to-end training in a simulated environment with a virtual user \cite{amrehn17uinet,du23click}. These virtual users mimic actions of human users in an automated manner, which is sufficient for model training and evaluation. Some IS frameworks incorporate multimodal inputs, including text prompts \cite{zhao25biomedparse} or diverse prompt mixtures \cite{liu24segnext}. Click- and scribble annotations serve as weak labels for guiding IS model predictions \cite{wong24scribbleprompt,lin16scribbleseg}. Click-based annotations have also been used for region of interest selection in transfer function design \cite{engel25transfer}. Annotation platforms \cite{diaz24monai} can be leveraged to annotate large-scale datasets for IS \cite{cheng24ISdata}, in turn leading to coherent and high-quality ML models. Powerful IS systems like SAM-Med2D \cite{cheng23sammed2d}, MedSAM \cite{ma24}, \nninter\ \cite{isensee25nninter}, \vista\ \cite{he24vista3d}, and others \cite{yuxin24segvol,gao25showandsegment,rokuss25lesionlocator,wang25sammed3d,Ma25MedSAM2} have been developed for the medical domain. Recently, memory-attention modules \cite{orbesarteaga25} have been introduced to improve IS performance by storing past user inputs and segmentation states to enable temporal context integration.

\textbf{Comparison to \meow:} In contrast to monolithic IS models that are trained end-to-end using a fixed interaction strategy, \meow\ is modular and thus flexible. Processing is handled by combining a frozen foundation model \cite{bommasani21foundation} as feature extractor with a transduction module, supporting diverse user interactions and different annotation styles. \meow\ makes predictions by combining nearest neighbor retrieval \cite{jegou11product,johnson21NN,douze24faiss} and correspondence matching \cite{lowe04sift,sun21loftr,edstedt24roma,mariotti24spherical} with transductive reasoning \cite{vapnik98statistical,zhou03transduction}. Hierarchical message passing \cite{pearl82mp,talbot06flow,frey07affinity,givoni11hierarchical} is applied as a post-processing step to ensure spatial consistency. We provide qualitative comparisons of our framework w.r.t. different baselines and semantic segmentation methods in \cref{tab:qualitative_comparison}.

\subsection{Open-Set and Open-World Recognition}
Closed-set recognition involves processing known structures that a model has been trained on. Open-set (OS) recognition involves identifying or rejecting unknown structures \cite{geng21opensetsurvey,nunes23survey}. Generalized OS recognition involves not only identifying or rejecting unknowns, but also being able to make predictions for them \cite{hong24GOSS,liu25unifiedopen}. Open-world (OW) recognition \cite{scheirer13OW,bendale15openworld,cruz25OW_problems} goes even further by incorporating unknown structures into the knowledge base during deployment. This requires robust novelty detection \cite{kim25OWCL} and is closely related to continual learning \cite{wang24survey,gonzalez23lifelong}, where models accumulate knowledge over time. Although test-time adaptation \cite{xiao24tta} also aims at improving performance on unknown structures during deployment \cite{schoen24adaptingSAM}, it does not integrate the continual learning aspect of OW recognition. This paradigm has gained traction in classification \cite{li24OWclass,li25OWclass}, object detection \cite{joseph21openworldobj,dong24OWobject}, semantic segmentation \cite{sodano24openworld,li26wowseg}, instance segmentation \cite{kalluri23openworld}, and panoptic segmentation \cite{sodano24panoptic}. However, interactive segmentation has not been studied in this context. Medical IS foundation models \cite{he24vista3d,zhang25medsegx} address OS segmentation of novel classes via interactive branches and learned context- and prompt encoders. While some methods improve models on known classes \cite{huang25SAM} or specific queries during deployment \cite{wang18,chen25tta}, they often fail under data distribution shifts \cite{zhou23OW_datashiftdegrade,rafi26OSshift,dong25OSadaptation}. Moreover, they often suffer from data-hungry and inefficient test-time fine-tuning. This creates high computational overhead that is prohibitive for clinical applications, where data-scarce environments demand low-latency, deterministic responses \cite{phuntsho25limitations}.

\textbf{Comparison to \meow:} As a memory-based, transductive reasoning framework, our approach is able to perform OW-IS via decoupling processing from knowledge \cite{weston15memory,graves16}. The latter is offloaded to a memory module containing labeled data, effectively making it a memory-augmented architecture \cite{krotov16memory,ramsauer21hopfield,lewis20RAG}. In contrast to monolithic models that rely on learned decoders \cite{he25fewshot}, this architectural choice aligns with recent findings that pre-training on procedural data \cite{munoz25procedural} is effective for decoupling perception from task-specific knowledge. The overall approach results in a flexible, composable, and scalable design which is suited for OW-IS. Learning capabilities include fast learning, transfer learning, few-shot learning, continual learning, and unlearning.

\begin{table}[t]
\scriptsize
\centering
\caption{Qualitative comparison of \meow\ to baselines and semantic segmentation methods.}
\begin{tabular}{lp{1cm}p{1cm}ccp{1cm}}
\hline
  \textbf{Method} & \textbf{modular / flexible} & \textbf{interaction- agnostic} & \textbf{OS} & \textbf{OW} & \textbf{real-time learning}\\
  \hline
  \suprem \cite{li24unetsuprem} & \xmark & \none & \xmark & \xmark & \xmark\\
  \totseg \cite{wasserthal23totseg} & \xmark & \none & \xmark & \xmark & \xmark\\
  \parm \cite{winwim24} & \cmark & \none & \xmark & \xmark & \cmark\\
  \sammed \cite{wang25sammed3d} & \xmark & \xmark & \cmark & \xmark & \xmark\\
  \vista \cite{he24vista3d} & \xmark & \xmark & \cmark & \xmark & \xmark\\
  \nninter \cite{isensee25nninter} & \xmark & \cmark & \cmark & \xmark & \xmark\\
  \meow & \cmark & \cmark & \cmark & \cmark & \cmark\\
\end{tabular}
\label{tab:qualitative_comparison}
\end{table}

\section{Method}
We begin with a problem formulation for open-world interactive segmentation (OW-IS) in \ref{method:problem}, followed by a brief overview of our solution from a user's perspective in \ref{method:meow_overview}. In \ref{method:elements_of_meow} and \ref{method:IS_pipeline}, we provide details about our approach to IS. In \ref{method:openworld}, we describe open-set and open-world capabilities of our approach. Lastly, we propose some general improvements to the original \parm\ method in \ref{method:improving_parmesan}. A list of symbols and abbreviations is given in Appendix V.

\subsection{Problem Formulation}
\label{method:problem}

\textbf{Interactive Segmentation (IS):} In 3D interactive segmentation, a user $\mathcal{U}$ actively guides an interactive segmentation model $\mathcal{S}$ to segment a structure of interest (SoI). The overall objective of IS is maximizing segmentation quality w.r.t. an SoI while at the same time minimizing the required user guidance (e.g., number and complexity of interactions). Moreover, an IS system must operate at low latency to enable meaningful interactions for $\mathcal{U}$.

\textbf{Open-World Interactive Segmentation (OW-IS):} In OW-IS, $\mathcal{S}$ is capable of segmenting an SoI while also learning from own predictions. Segmentation not only includes known classes (closed-set IS), but extends to unknown classes and domains (open-set IS). The overall objective aligns with the objective for IS, but also requires learning from classes and domains that are unknown to $\mathcal{S}$. In addition to low-latency interactions, an OW-IS system should preferably learn (and unlearn) at low latency in order to adapt fast.

Grounded in mixed-initiative principles \cite{horvitz99}, effective collaboration between $\mathcal{U}$ and $\mathcal{S}$ requires leveraging automated predictions when confidence is high while relying on user guidance when confidence is low. Designing a system to balance both aspects is critical for enabling efficient interaction.

We consider structures of interest (SoIs) to be flexible target regions that do not need to correspond to predefined semantic classes. They may contain holes and can consist of multiple disconnected or nested parts. The goal of $\mathcal{U}$ is to segment these SoIs by interacting with $\mathcal{S}$. Considering an SoI with its ground truth binary segmentation mask (i.e., ground truth label), any additional information such as its class, semantic description, input domain, etc. are considered metadata. Depending on $\mathcal{S}$'s knowledge about an SoI and its metadata, four different class categories can be defined in the OW setting \cite{geng21opensetsurvey}:
\begin{itemize}
    \item KKCs ("known-known"): Classes with unambiguous labels and semantic / attribute metadata.
    \item KUCs ("known-unknown"): Classes explicitly labeled as unknown (e.g., background or any unknown structure).
    \item UKCs ("unknown-known"): Classes with no samples at training, but available semantic / attribute metadata.
    \item UUCs ("unknown-unknown"): Classes with neither data at training, nor metadata available.
\end{itemize}

\subsection{Overview of \meow\ for OW-IS}
\label{method:meow_overview}

When presented with a volumetric image (e.g., a CT or MRI scan), a user $\mathcal{U}$ first selects an SoI by clicking on it. A local region of fixed resolution (e.g., $128\times128\times128$ voxels) is subsequently cropped around the selection point, yielding a query image $x_q$ that is centered on a portion of the SoI.

The query is then processed to make a first prediction using the initial selection point as a weak positive label (see \cref{fig:init_phase_plot}). After this initialization phase, the alternating interaction phase starts, where $\mathcal{U}$ guides the model to modify and improve its prediction. At every interaction step, $\mathcal{U}$ provides guidance and $\mathcal{S}$ makes a prediction, which is then visually displayed to $\mathcal{U}$ for further refinement. Guidance is provided by placing additional (positive or negative) clicks and brushes, or by modifying interactive hyperparameters via sliders. $\mathcal{U}$ may end interaction when a certain segmentation quality is reached.

The final result may then be learned depending on the use case. If another, similar query is encountered in the future, previously learned samples can be leveraged to improve OW-IS objectives. Learning with \meow\ is fully reversible and previously learned samples can easily be unlearned if they are no longer required.

\subsection{Elements of \meow}
\label{method:elements_of_meow}

\begin{figure}[t]
	\centering
	\includegraphics[width=1.\linewidth]{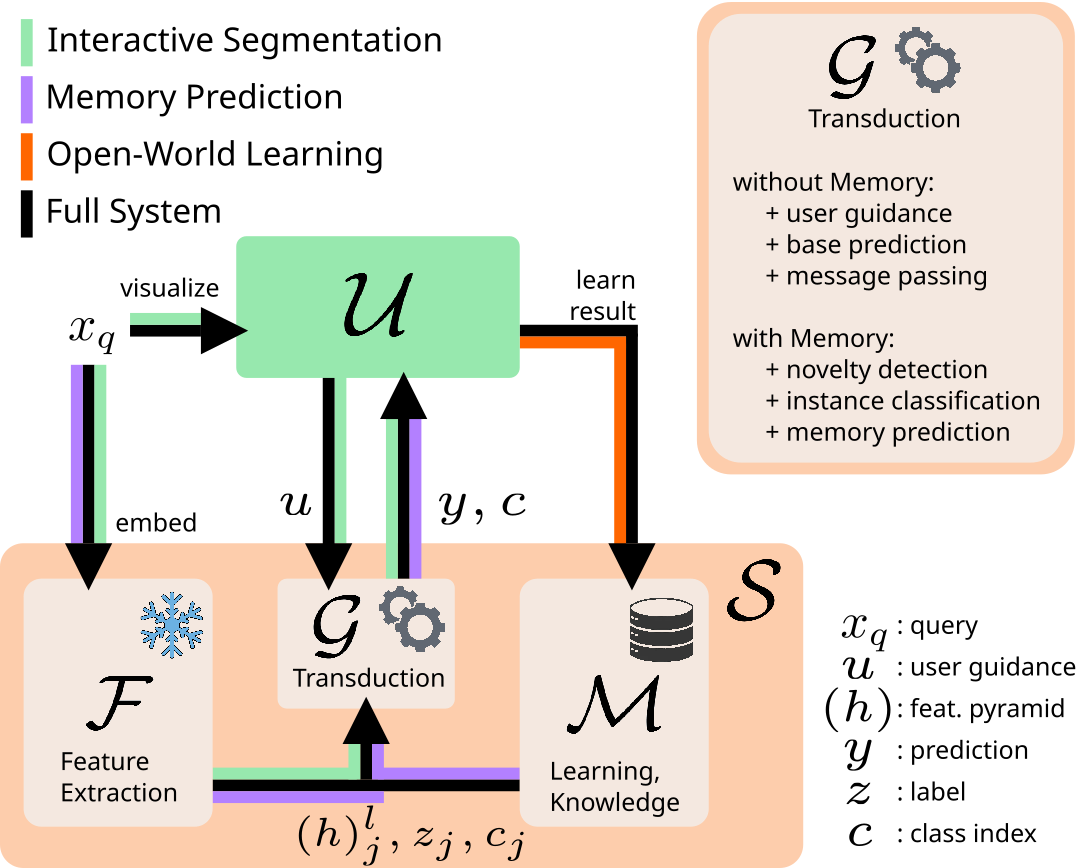}
	\caption{Data flow diagram of our OW-IS approach \meow: A memory module $\mathcal{M}$ accumulates knowledge over time and is used in combination with a frozen feature extractor $\mathcal{F}$ and a parameter-free transduction module $\mathcal{G}$ for making predictions. A user $\mathcal{U}$ interacts with the system by providing guidance to improve results. Final results may be learned via memory consolidation. Learned samples can instantly be leveraged to improve performance for future predictions.}
	\label{fig:open-world_skizze}
\end{figure}

We define an IS system as the combination of an interactive segmentation model $\mathcal{S}$ and a user $\mathcal{U}$ (see \cref{fig:open-world_skizze}). Given a query image $x_q$ along with some user guidance, $\mathcal{S}$ makes segmentation predictions for an SoI within $x_q$ (see \cref{fig:interactionplot}).

For data, we assume regular Cartesian voxel grids (i.e., uniform structured rectilinear grids) on uniformly discretized Euclidean domains. This naturally accommodates diverse spatiotemporal data modalities, including volumetric medical images (CT, MRI), videos, and Eulerian simulation grids in numerical modeling. Let $\Omega^l=(N_1^l \times N_2^l \ldots N_d^l)$ denote the $d$-dimensional spatial domain of a representation at resolution-level $l=1,\ldots,L$, with $\Omega^1$ having the highest resolution and $\Omega^L$ having the lowest resolution. Let $x \in X \subseteq \mathbb{R}^{\Omega^1}$ denote a model input. For example, 3D CT data has $\Omega^1 = (H^1 \times W^1 \times D^1)$, where $H$, $W$, and $D$ indicate grid-resolutions in height, width, and depth.

\textbf{User $\mathcal{U}$:} For guiding predictions, $\mathcal{U}$ relies on a cognitive mental model of an SoI's outline, which we formalize as the ground truth binary segmentation mask $z_{gt} \in \{0, 1\}^{\Omega^1}$. $\mathcal{U}$ might also have access to an SoI's metadata like semantic- and attribute information, which may be used by $\mathcal{S}$ (details in \cref{method:openworld}). In the context of this work, $\mathcal{U}$ is typically a medical expert with domain expertise (e.g., a radiologist). At inference, $\mathcal{U}$ and $\mathcal{S}$ interact in an alternating manner to segment an SoI. Model predictions $y \in [0, 1]^{\Omega^1}$ can be thresholded into a pseudo-label binary mask $z_p = \mathbb{I}(y \geq \tau)$ with $z_p \in \{0, 1\}^{\Omega^1}$ and $\tau \in \mathbb{R}$ being the classification threshold. These masks can then be learned via memory consolidation to improve future predictions. While $\mathcal{U}$'s role is providing weak labels and hyperparameter changes for guidance (see \cref{fig:init_phase_plot}), $\mathcal{S}$ provides corresponding segmentation predictions $y$.

For adapting to specific use cases and diverse annotation preferences of different users, we introduce several components to increase overall flexibility of our method. These components can be controlled independently and dynamically (e.g., via sliders) as interactive hyperparameters $\boldsymbol{\iota}$, allowing $\mathcal{U}$ fine-grained control over segmentation behavior. Depending on the technical expertise of $\mathcal{U}$, they may prefer to bypass fine-grained control via $\boldsymbol{\iota}$, falling back to default values typically obtained through automated hyperparameter optimization (see Appendix IV for details).

\textbf{Segmentation Model $\mathcal{S}$:} In our framework, the interactive segmentation model $\mathcal{S}$ consists of three modules: a frozen feature extractor $\mathcal{F}$, a memory module $\mathcal{M}$ for learning, and a parameter-free transduction module $\mathcal{G}$ for making predictions (see \cref{fig:eye_catcher} and \cref{fig:open-world_skizze}). We attach $\mathcal{G}$ to $\mathcal{F}$ to form a functional IS system. The system is further enhanced with a memory module $\mathcal{M}$ to enable OW-learning capabilities. We now introduce these modules in more detail and describe how they operate.

Let $z^\pm \in \{+1, -1, \varnothing \}^{\Omega^1}$ denote a weak user label, representing the combination of positive foreground ($+1$), negative background ($-1$), or no label ($\varnothing$) w.r.t. an SoI at image resolution $\Omega^1$ (i.e., a trinary mask). We view $z^\pm$ as a special type of sparse labels where some voxels are left unlabeled. Weak labels may be composed of any number of clicks and brushes in any spatial direction, including off-axis directions. The combination of all clicks and brushes is $z^\pm$. Brushing can be done in both 2D (via circles, lassos, etc.) or 3D (via spheres, cubes, etc.). Unlabeled voxels are being ignored for processing. Weak labels are combined with a set of interactive hyperparameters $\boldsymbol{\iota}$, which control different aspects of how predictions are calculated by $\mathcal{G}$ (details in Appendix IV). Together, they form a unified user guidance $u=\{z^\pm, \boldsymbol{\iota}\}$. During the alternating interaction phase, $u$ is partially or fully modified by $\mathcal{U}$ in-between calls to $\mathcal{G}$.

\textbf{Feature Extractor $\mathcal{F}$:} A pre-trained feature extractor $\mathcal{F}$ with frozen weights $\theta$ extracts a multi-scale feature pyramid $(h)^l = \mathcal{F}(x)$, yielding hidden representations $h^l \in \mathbb{R}^{C_l \times \Omega^l}$ at each resolution-level $l=1,\ldots,L$, where $C_l$ denotes the number of feature channels. The number of levels $L$ depends on $\Omega^1$, scaling down by a factor of two at each level until reaching the lowest resolution $\Omega^L\le(2\times2\times2)$. In case of incomplete feature pyramids, missing levels are interpolated or extrapolated from existing representations. Because $\mathcal{F}(x)$ is independent of user guidance $u$, feature extraction is computed only once during the initialization phase (details in \cref{method:IS_pipeline}).

Our requirement for $\mathcal{F}$ is that it is capable of providing feature pyramids $(h)^l$, which is generally true for hourglass-shaped neural network architectures like the U-Net family \cite{ronnberger15unet,zhao17pspnet,isensee21nnunet}. However, we do not assume that $\mathcal{F}$ is pre-trained specifically for the task of IS. Instead, we can utilize models trained for other tasks such as semantic segmentation, instance segmentation, panoptic segmentation, or regression. For some use cases, $\mathcal{F}$ may even be completely untrained.

\textbf{Memory Module $\mathcal{M}$:} A memory module $\mathcal{M}$ stores \mbox{$j=1,\ldots,m$} labeled samples $\mathcal{M} = \bigl\{ \bigl(x, z, c, (h)^l\bigr)_j \bigr\}$, where \mbox{$c_j \in \{0,\ldots,K\}$} denotes the semantic class of the corresponding SoI. We place no restrictions on the class definition, allowing $c$ to group any anatomical, pathological, operational, or custom semantic concept. The process of preparing and writing samples to $\mathcal{M}$ is referred to as \textit{learning through memory consolidation}. Consolidation strategies may involve quantization and compression in latent- and spatial dimensions for efficient storage. Our strategy involves local novelty search to find and discard voxels in $(h)^l$ that are redundant w.r.t. other voxels \cite{winwim24}. Alternatively, one might discard voxels simply by random sampling. Making $0<L_{sp}\le L$ levels sparse reduces the total number of voxels $p = \prod_d\Omega^l$ of level $l$ by a factor of $2^{d\cdot \max(L_{sp}-l+1,0)}$. This results in resolutions $\Omega^l$ being identical across all sparse levels. We further apply compression of $(h)^l$ in the channel dimension (more details in \cref{method:improving_parmesan}). 

We distinguish between a long-term memory module $\mathcal{M}$, which accumulates and stores episodic memories, and a working memory $M$, which is used more flexibly during a forward pass for making predictions. Depending on the processing step of the inference pipeline, $M$ may contain the full $\mathcal{M}$, a subset of $\mathcal{M}$, or the current query itself (details in \cref{method:IS_pipeline} and \cref{method:openworld}). Fast learning and unlearning of individual samples is achieved simply through write and delete operations on $\mathcal{M}$.

\textbf{Transduction Module $\mathcal{G}$:} Given a query feature pyramid $(h)_q^l$ along with user guidance $u$ and some working memory $M$, a transduction module $\mathcal{G}$ makes segmentation predictions \mbox{$y=\mathcal{G}(\mathcal{F}(x),~u;~M)$} (see \cref{fig:open-world_skizze} and \cref{fig:IS_skizze}). $\mathcal{G}$ essentially replaces learned decoders in a parameter-free manner. The process of leveraging stored samples for making predictions is referred to as \textit{transductive reasoning}. To obtain a prediction $y$, transductive reasoning is applied w.r.t. user guidance $u$ (for interpreting weak labels $z^\pm$), $\mathcal{M}$ (for making memory a prediction $y_\mathcal{M}$), as well as $x_q$ itself (for post-processing raw predictions $y_r$).

$\mathcal{G}$ performs approximate nearest neighbor (NN) search \cite{indyk98approxNN} for non-parametric context-retrieval and reasoning (also see Appendix I). Operating over multi-scale feature pyramids, this mechanism enables highly scalable and parallelized correspondence matching suited for various vision tasks \cite{winwim24}. For every query voxel, NN search is performed against two kinds of memory contents: unpaired contents from stored samples and paired contents from the query itself. This yields semantic affinities that are used in subsequent reasoning steps. Comparable to retrieval-augmented generation in language models \cite{lewis20RAG}, our approach first retrieves relevant information via similarity-based NN search, followed by reasoning over them to obtain predictions. Post-processing via hierarchical message passing (HMP) yields smooth and self-consistent segmentations (details in \cref{method:IS_pipeline}).

While $\mathcal{G}$ is primarily responsible for segmentation predictions from both user guidance and memory, it also comprises functions relevant for learning, novelty detection, and instance classification (details in \cref{method:openworld}).

\subsection{Interactive Segmentation Pipeline}
\label{method:IS_pipeline}

\begin{figure}[t]
	\centering
	\includegraphics[width=1.\linewidth]{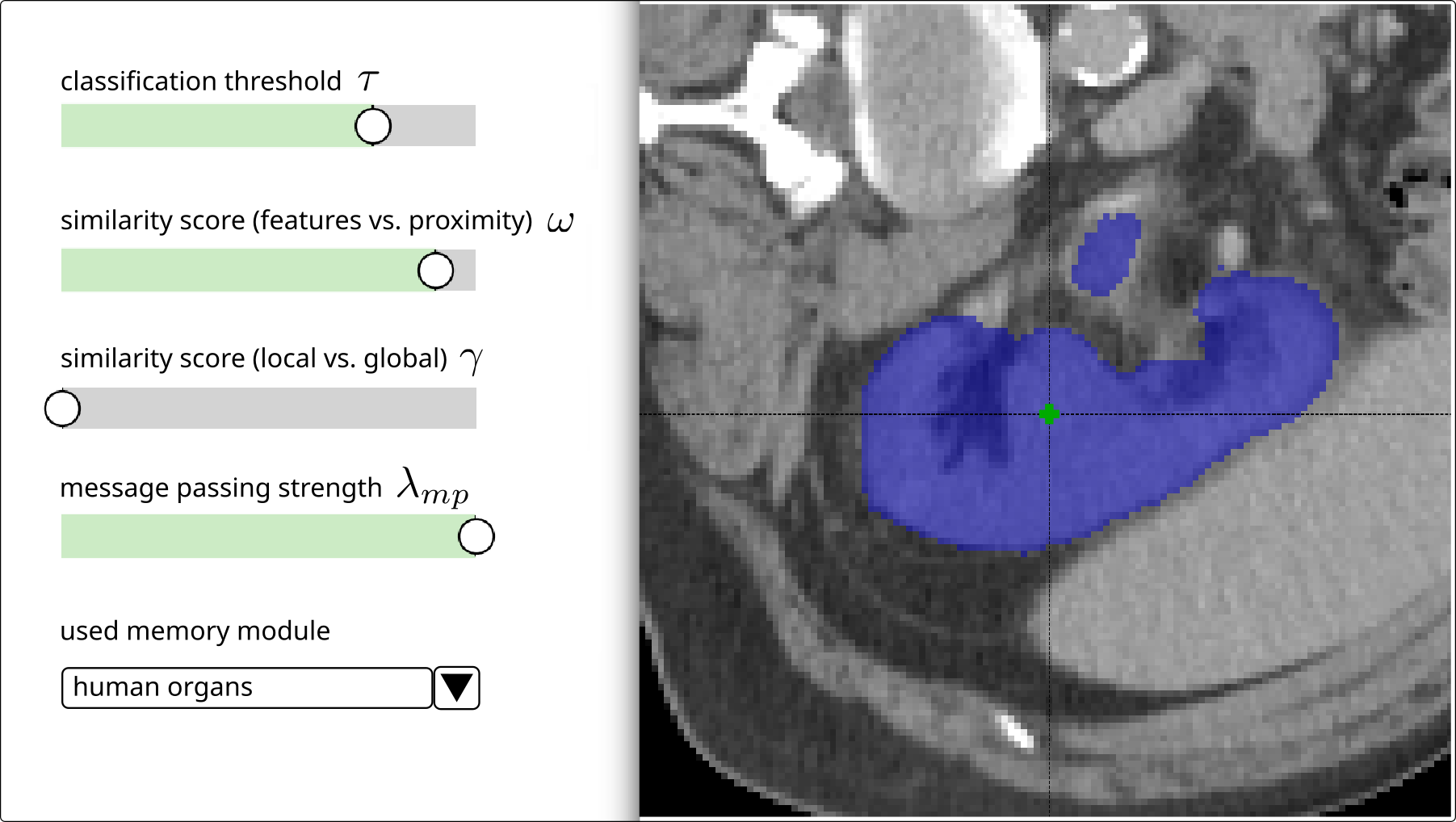}
	\caption{After the initialization phase, the alternating interaction phase starts by presenting several visual components to $\mathcal{U}$: the query image $x_q$ with an overlay of the initial prediction $y$ (in blue), as well as the weak label $z^+$ at the center (in green). A crosshair indicates the current cursor position on the canvas. $\mathcal{U}$ can navigate through the volume via the mouse wheel or via keyboard buttons. On a control panel, interactive hyperparameters $\boldsymbol{\iota}$ can be adjusted via sliders. A memory module suited for the task at hand can be selected. After every user action (i.e., clicks, brushes, or hyperparameter changes), $\mathcal{G}$ is called to update the prediction.}
	\label{fig:init_phase_plot}
\end{figure}

For IS using \meow, inference is divided into an initialization phase and an alternating interaction phase. An additional learning step is present in an open-world scenario (see \cref{method:openworld}). \textbf{The initialization phase} is executed once before alternating interactions start. This phase covers all computations that are independent from user guidance, including:
\begin{enumerate}
    \item Embedding: A query $x_q$ is embedded into $(h)_q^l=\mathcal{F}(x_q)$ with respective resolutions $\Omega^l$.
    \item Feature Compression Weights: Linear projection weights for compressing $(h)^l$ in the channel dimension are calculated via linear autoencoders (details in \cref{method:improving_parmesan}).
    \item Initialize Hierarchical Message Passing (HMP): A query-specific HMP graph is calculated by finding NNs between voxels of $(h)_q^l$ for all $l=1,\ldots,L_{mp}\le L$ levels, utilizing \meow's NN search capabilities (see Appendix I, Algorithm 1).
    \item Memory- and Initial Prediction: A memory prediction $y_\mathcal{M}$ may be made by utilizing stored memory samples (details in \cref{method:openworld}). An automated weak positive label $z^+$ is placed at the center of $x_q$ (i.e., where $\mathcal{U}$ first clicked into the SoI), resulting in a base prediction $y_b$. It is merged with $y_\mathcal{M}$ to make an initial prediction $y$.
    \item Setup Interaction: The alternating interaction phase with $\mathcal{U}$ begins through a graphical interface, where a query image $x_q$, weak labels $z^\pm$, the initial prediction $y$, and interactive hyperparameters are visualized (see \cref{fig:init_phase_plot}).
\end{enumerate}

\textbf{The alternating interaction phase} covers all computations that depend on user guidance, i.e., correcting and refining model predictions. The dynamics of this phase are driven by alternating between user actions and calls to $\mathcal{G}$. Based on visual overlays of model predictions $y$, $\mathcal{U}$ interacts with the system by placing weak labels $z^\pm$ (i.e., clicking and brushing) and adjusting interactive hyperparameters $\boldsymbol{\iota}$. Weak labels $z^\pm$ denote the cumulative aggregation of all positive and negative clicks and brushes across all interaction steps. Clicking and brushing tools can also be used to erase previously drawn labels.

\begin{figure*}[t]
\begin{minipage}{0.6\textwidth}
    \includegraphics[width=\textwidth]{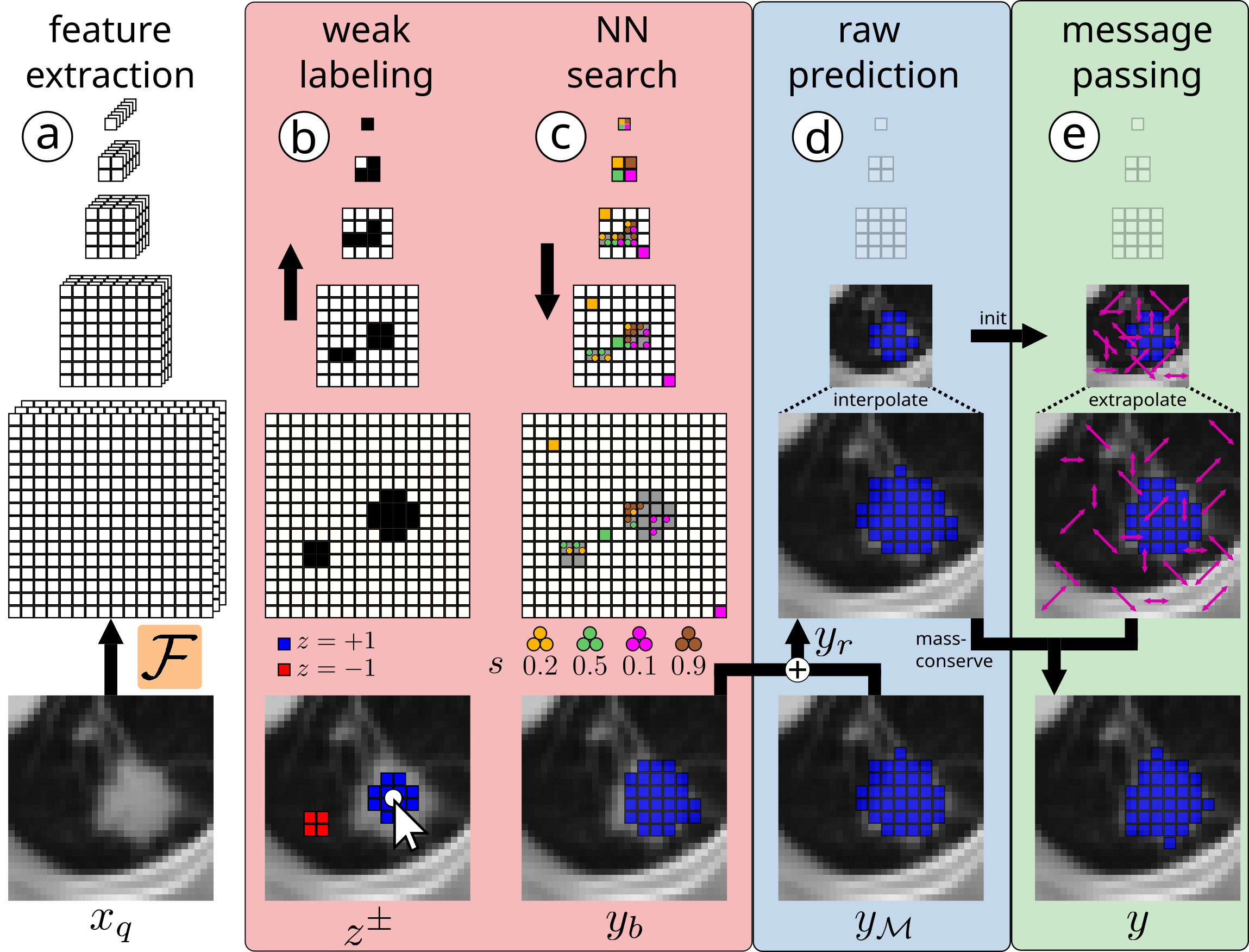}
  \end{minipage}
  \hfill
  \begin{minipage}{0.39\textwidth}
    \caption{Interactive segmentation with \meow. The three processing steps of a call to $\mathcal{G}$ include an interactive base prediction (red), optional merging of a memory prediction (blue), and hierarchical message passing (green). a) A feature pyramid $(h)_q^l$ is calculated once by applying a feature extractor $\mathcal{F}$ to a query input $x_q$. b) Weak user labels $z^\pm$ for model guidance (positive: blue, negative: red) determine labeled parts $(\mu)_q^l$ of $(h)_q^l$. c) Approximate, hierarchical $k$-NN w.r.t. $(\mu)_q^l$ illustrated for 4 exemplary query voxels (in different colors), followed by label retrieval and reduction to obtain a base prediction $y_b$. d) Combining $y_b$ with an optional memory prediction $y_\mathcal{M}$ yields a raw prediction $y_r$. e) Hierarchical message passing for smooth and consistent segmentations.}
    \label{fig:IS_skizze}
  \end{minipage}
\end{figure*}

We now describe how $\mathcal{G}$ handles weak user labels $z^\pm$, i.e., labels without complete, voxel-wise annotations. Given the feature pyramid $(h)_q^l$ and user guidance $u$, predictions $y$ are made in three processing steps (also see \cref{fig:IS_skizze}):
\begin{enumerate}
    \item interactive base prediction $y_b$,
    \item optional merging of memory prediction $y_\mathcal{M}$,
    \item and hierarchical message passing.
\end{enumerate}
Predictions $y$ are calculated from scratch with every call of $\mathcal{G}$.

In \textbf{processing step \#1} (Appendix I, Algorithm 1), an interactive base prediction $y_b$ is made by $\mathcal{G}$ to incorporate $u$ (\cref{fig:IS_skizze} b). We begin by filling a working memory $M$ with the query itself, along with weak labels $z^\pm$. To indicate labeled regions within $(h)_q^l$, a pyramid binary mask $(\mu)_q^l$ is constructed by resizing $|z^\pm|$ via linear interpolation to resolutions $\Omega^l$, followed by non-zero thresholding. Fast approximate, hierarchical $k$-NN search is then performed for every voxel in $x_q$, restricted to the sparsely labeled regions $(\mu)_q^l$. Starting from the global representation $h^L$, levels are traversed downward through the feature pyramids, thereby retrieving both globally and locally similar stored voxels in $M$ (\cref{fig:IS_skizze} c). Since resolutions $\Omega^l$ increase $2^d$-fold from level $l$ to $l-1$, $k$ is reduced accordingly such that computational complexity remains quasi-constant throughout the hierarchical search process \cite{winwim24}.

For any level $l$, voxel-wise cosine similarities between query features and memory features are 
\begin{align}
s_{cos}^l(h_q^l, h_j^l) = \frac{h_q^l \cdot h_j^l}{\lVert h_q^l\rVert \, \lVert h_j^l\rVert} \in [-1, 1]^{\Omega^l},
\end{align}
with feature-similarities being
\begin{align}
s_f^l(h_q^l, h_j^l)=\frac{1}{2}\left(1+s_{cos}^l(h_q^l,h_j^l)\right) \in [0, 1]^{\Omega^l}.
\end{align}
To avoid ambiguous matches originating from symmetries and repeated patterns \cite{mariotti24symmetries}, we employ a combination of both feature-based and proximity-based scoring for memory search as a geometric prior. For any NN of a query voxel in level $l$, accumulating similarities over levels yields 
\begin{align}
    s^l=\prod_{i=l}^L (\omega s_f^i + (1 - \omega) s_o^i)\in [0, 1]^{\Omega^l},    
\end{align}
where $s_o$ are proximity-similarities calculated using Euclidean distances between query positions and memory positions on the voxel grid. The product over levels acts as a momentum component that integrates the search trajectory into $s^l$. The hyperparameter $\omega \in [0, 1]$ controls the contribution of $s_f$ and $s_o$ to the final similarity score $s^l$ of level $l$.

NN search progresses from global, abstract representations to local, primitive representations to find semantically meaningful matches. Traversing through levels increases spatial resolutions, necessitating a reciprocal reduction in $k$ (controlled by $k_{start}$ and reduction factor $\phi$) to maintain near-constant hardware overhead \cite{winwim24}.

After finding $k$ NNs for every query voxel, we retrieve respective final similarity scores $s_j^1$ with $j=1,\ldots,k$ along with the corresponding label values of $z^\pm$. A base prediction $y_b$ is computed via weighted reduction over the top-$k$ NNs. When $(\mu)_q^1$ are extremely sparse (e.g., by clicking only a handful of voxels), distant NNs that are dissimilar w.r.t. other NNs can introduce noise for naive weighted averaging. To mitigate this, we adopt a continuous Hopfield layer \cite{ramsauer21hopfield} without learnable parameters, where a hyperparameter $\beta_{nn}$ influences how strongly reduction weights $w_j$ are committed towards higher similarity scores.

The \textbf{processing step \#2} is optional depending on the availability of a memory prediction $y_\mathcal{M}$. Given a working memory $M$ containing labeled samples (see \cref{eq:mem}), a dense prediction $y_\mathcal{M}$ is made by finding and retrieving top-$k$ NNs from $M$ for every query voxel. Retrieved labels are reduced via $y_\mathcal{M}=\sum_k \text{softmax}_k(s_j) \cdot z_j$. Here, $s_j$ denotes the similarity score for the $j$-th retrieved NN voxel. We integrate $y_\mathcal{M}$ into the main forward pass by fusing it with the base prediction $y_b$ (\cref{fig:IS_skizze} d, also see Appendix I, Algorithm 2). The same fusion approach applies to other scenarios, e.g., when continuing a previous IS session or when fusing predictions from multiple different methods. In our framework, the calculation of $y_\mathcal{M}$ is part of the initialization phase preceding user interaction. A memory prediction is therefore only computed once and remains fixed during user interaction.

In \textbf{processing step \#3}, raw predictions $y_r$ are refined via hierarchical message passing (HMP, also see Appendix I, Algorithm 3). Recall that $y_r$ is composed of independent predictions for every query voxel, thereby neglecting spatial priors like local smoothness and boundary awareness. By capturing intra-query correlations only indirectly through feature embeddings, $y_r$ is prone to metrication artifacts \cite{kolmogorov03graphcuts} (e.g., blockiness) and imprecise segment boundaries. These limitations are resolved by exploiting intra-query correlations via HMP throughout $(h)_q^l$ for levels $l=1, \ldots, L_{mp}\leq L$ (\cref{fig:IS_skizze} e). The additional use of $L_{mp}>1$ low-res representations leads to smoother and more consistent predictions w.r.t. $L_{mp}=1$ (i.e., \parm's original approach), as well as being capable of correcting large metrication artifacts and perturbations. HMP employs a coarse-to-fine refinement strategy w.r.t. spatial resolutions, where higher levels apply global modifications and lower levels apply local modifications. As in processing step \#1, feature pyramid levels are traversed downward from $l=L_{mp}$ to $l=1$.

In our inference pipeline, a static HMP graph is calculated during the initialization phase leveraging \meow's $k$-NN search capabilities. Given $(h)_q^l$ both as query and working memory content, we find $\kappa$ NNs for every query voxel within the query itself. Consequently, the top-1 NN is the respective query voxel itself. Search is dense where $\forall \mu_i \in (\mu)_q^l, \mu_i = 1$ (i.e., all voxels labeled). For calculating HMP graph edges $(e)_j^l$, NN indices $(i)_j^l$ with similarity scores $(s)_j^l$ are returned instead of label-retrieval and reduction (also see Appendix I, Algorithm 1).

For every HMP level, node features $y$ and weak labels $z^\pm$ are first interpolated to the respective resolution $\Omega^l$. Then, $\eta_{mp}$ message passing steps are taken while keeping weakly labeled nodes fixed. Update strength $\lambda_{mp}$ origins from graph neural network theory \cite{li16gnn} and can be controlled interactively by $\mathcal{U}$.

Through message passing, the over-smoothing problem \cite{rusch23} tends to make node features more similar. This can lead to vanishing segments, especially when SoIs are small. We mitigate this effect by combining artificial sharpening with another Hopfield layer, as well as applying mass-conservation in terms of foreground voxels. Sharpening strength around the classification threshold $\tau$ is controlled by $\sigma_{mp}$. A continuous Hopfield layer $(e)_j^l=\text{softmax}_\kappa(\beta_{mp} \cdot (s)_j^l)$ without learnable parameters is controlled by a hyperparameter $\beta_{mp}$, which determines how strongly graph edges $(e)_j^l$ are committed towards more similar NNs. Larger values of $\beta_{mp}$ effectively make the HMP graph more sparse.

Retaining certain quantities in the form of conservation laws is beneficial for solving problems related to dynamical systems \cite{greydanus19HNN,chen20symplectic,hoedt21mclstm,rodriguez21,cranmer21lagrangian}. To further mitigate over-smoothing, we employ mass-conservation in terms of voxel count by adjusting $\tau$ such that the number of foreground voxels remains constant w.r.t. raw prediction $y_r$. HMP dynamics can therefore only re-distribute mass without growing or shrinking the resulting segment.

\subsection{Open-World Learning with \meow}
\label{method:openworld}

We now focus on introducing \meow\ as an open-world interactive segmentation (OW-IS) system, incrementally learning from its own predictions during deployment under user guidance. Building upon its flexible learning and inference framework, \meow\ is uniquely suited for OW-IS. By extending the base architecture ($\mathcal{F}\cup\mathcal{G}$) with a memory module $\mathcal{M}$, our system adapts and learns from novel structures via memory consolidation (see \cref{fig:open-world_skizze}). Through this setup, \meow\ supports multiple learning paradigms: incremental, continual, transfer, out-of-distribution, few-shot, and OW learning. Furthermore, our model allows fast learning and unlearning of individual samples at scale, as well as performing novelty detection (ND) and instance classification (IC).

\textbf{Known vs. Unknown:} Given the four OW class categories $\text{KKCs}$, $\text{KUCs}$, $\text{UKCs}$, and $\text{UUCs}$ as defined in \cref{method:problem}, we further refine these categories based on their status relative to the backbone $\mathcal{F}$ (via training), the memory $\mathcal{M}$ (via consolidation), or both, yielding twelve distinct combinations. Since we keep learnable parameters $\theta$ frozen, the "known" status w.r.t. $\mathcal{F}$ is governed by the scope of the pre-training data. Note that replacing or fine-tuning $\mathcal{F}$ necessitates re-embedding existing memory samples to maintain consistency across feature pyramids.

For the task of IS, $z_{gt}$ are ground truth binary segmentation masks, whereas ground truth class indices $c_{gt}$ and associated semantic information are treated as metadata. Our framework does not strictly require $\mathcal{U}$ to provide $c_{gt}$ due to its ND and IC capabilities. Consequently, our method supports all twelve OW category combinations, as well as flexibly modifying and merging them throughout a model's operational lifecycle.

\textbf{Learning with \meow:} In our framework, an OW-learning system is initialized with an empty memory, where any incoming query is novel to $\mathcal{M}$ (i.e., an $\text{UKC}_{\mathcal{M}}$ or $\text{UUC}_{\mathcal{M}}$). The query gets assigned a class index $c\ge0$, e.g., by $\mathcal{U}$ providing the respective $c_{gt}$. While $c>0$ refer to known classes (i.e., $\text{KKC}_{\mathcal{M}}$), $c=0$ is regarded as the unknown class (i.e., $\text{KUC}_{\mathcal{M}}$).

An open-set (OS) segmentation result $y$ is obtained via IS (see \cref{method:IS_pipeline} and green pipeline in \cref{fig:open-world_skizze}). $\mathcal{U}$ can then judge segmentation quality and decide if the query and its prediction shall be learned (i.e., become a memory sample). A sample is learned by first creating a pseudo-label binary mask $z_p = \mathbb{I}(y \geq \tau)$ from the final prediction. The feature pyramid $(h)_q^l$ along with $x_q$, $z_p$, and $c$ are then stored in $\mathcal{M}$. It can accommodate storing arbitrary sets of ground truth labels $z_{gt}$ and pseudo-labels $z_p$, allowing hybrid approaches such as initializing $\mathcal{M}$ with some ground truth labeled samples while learning pseudo-labeled samples during deployment.

\textbf{Novelty Detection:} Assuming a non-empty $\mathcal{M}$, we apply ND and IC for predicting a class index $c$ and deciding on whether a segmentation prediction $y_\mathcal{M}$ from memory can be made. During the initialization phase, we calculate accumulated similarities $s_{f*,j}$ of small pyramid center-patches \mbox{$(h)_{q*}^l\subset(h)_q^l$} and \mbox{$(h)_{j*}^l\subset(h)_j^l$} between a query patch and $j=1,\ldots,m$ memory patches:
\begin{equation}
    s_{f*,j} = \prod_{i=1}^{L} s_{f*}^i(h_{q*}^i, h_{j*}^i).
\end{equation}
Let $\mathcal{N}_\mathsf{k}$ be the set of indices of the top-$\mathsf{k}$ most similar memory patches according to $\{s_{f*,j}\}$. 
The average similarity of the query to these neighbors is
\begin{equation}
    s_{f*}^q = \frac{1}{\mathsf{k}} \sum_{j \in \mathcal{N}_\mathsf{k}} s_{f*,j}.
\end{equation}
To get a robust estimate for typical intra-memory similarities $s_{f*}^M$, we average all pairwise similarities between memory patches in $\mathcal{N}_\mathsf{k}$. A query is considered novel if $\mathcal{M}$ is empty or if $s_{f*}^q < \zeta \cdot s_{f*}^M$ with the novelty threshold $\zeta\in[0,1]$ being a hyperparameter. If a query is novel, no memory prediction $y_\mathcal{M}$ is made and the query is considered unknown (i.e., associated with $c=0$).

\textbf{Instance Classification:} If a query is not novel, we perform IC by similarity-weighted voting over the class labels of the top-$\mathsf{k}$ NN patches to obtain a class prediction $c \ge 0$. This task corresponds to a multi-class IC problem with a variable number of classes. If $c>0$, the query is a $\text{KKC}_{\mathcal{M}}$ and being associated with the respective known class. If $c=0$, the query is a $\text{KUC}_{\mathcal{M}}$.

Given our focus on medical OW-IS, $\mathcal{U}$ typically has extensive domain expertise, including prior knowledge (i.e., semantic metadata) of the SoIs to be segmented (e.g., a specific pathology). In this case, $\mathcal{U}$ may verify the correctness of $c$ or correct it by providing $c_{gt}$. When $c_{gt}$ is provided upfront, ND and IC are skipped entirely. In scenarios where $\mathcal{U}$ can't provide $c_{gt}$ (i.e., $\text{UUCs}$), the predicted $c$ is used to continue.

\textbf{Memory Prediction:} A working memory $M$ is created by selecting $z$ and $(h)^l$ of all samples from $\mathcal{M}$ that either belong to the class $c$ or are among the top-$\mathsf{k}$ NNs obtained from IC:
\begin{equation}
\label{eq:mem}
    M = \bigl\{ \bigl(z, (h)^l\bigr)_i \in \mathcal{M}_* 
    \big| 
    \bigl((c_i = c) \land (c > 0)\bigr) 
    \lor 
    i \in \mathcal{N}_\mathsf{k} \bigr\},
\end{equation}

where $\mathcal{M}_* = \bigl\{ \bigl(z, (h)^l\bigr)_i \big| \bigl(x, z, c, (h)^l\bigr)_i \in \mathcal{M}\bigr\}$. If $M$ is empty, no memory prediction $y_\mathcal{M}$ can be made. If $M$ is non-empty, a dense prediction is made and integrated by fusing it with the base prediction $y_b$ as described in \cref{method:IS_pipeline}. Note that $y_\mathcal{M}$ can be calculated fully automatically without a user, thereby enabling method variants for OS and OW semantic or panoptic segmentation. While $M$ holds a single sample (namely the query itself) for making a base prediction $y_b$ in processing step \#1, $M$ may hold many samples for making a memory prediction $y_\mathcal{M}$.

Finally, $\mathcal{U}$ provides a binary decision (accept or reject) on whether to store the resulting segmentation $y$ in $\mathcal{M}$, thereby establishing an open-world learning loop that enhances predictive performance for subsequent queries.

\subsection{Additional Improvements to \parm}
\label{method:improving_parmesan}
Similar to \meow, \parm\ \cite{winwim24} is a transductive reasoning framework extending non-parametric classification \cite{nakata22} and kernel-based detection \cite{scholkopf98kernel,gauman05matchkernel} to dense prediction tasks (e.g., semantic segmentation or depth estimation) for continual learning. \parm\ makes non-interactive predictions using densely labeled, unpaired memory samples.

Starting from the original method, we introduce the following general improvements toward high-quality, fast, and efficient OW-IS for 3D data. These improvements are integral components of \meow\ and are applied in all configurations and experiments throughout this work:
\newline
\textbf{Extension to 3D:} We extend the method to 3D volumetric data such as CT and MRI scans. This includes using 3D feature pyramids from respective feature extractors, as well as updating $\mathcal{G}$ to efficiently operate on 3D grids.
\newline
\textbf{Feature Compression:} Storing raw feature pyramids $(h)^l$ requires substantial hardware memory resources, especially in 3D. To keep requirements low, \parm\ uses simple channel trimming to reduce the number of channels. In contrast, \meow\ uses linear projections for fast and efficient compression of $(h)^l$ in the channel dimension. Given one or more samples, optimal projection weights for every level $l$ are obtained through a linear autoencoder \cite{bourlard88linearAE} and singular value decomposition \cite{joreskog67pca,everitt84pca}. Compression weights are re-calculated for every incoming query and remain fixed once $\mathcal{M}$ is being used. If no data is available, projection weights may be randomly sampled instead \cite{bingham01random}.
\newline
\textbf{Kernels:} For $k$-NN search between query pyramid voxels and memory contents, we discard the originally proposed box-shaped search kernel in favor of two new kernels, the \textit{cross}-kernel and the \textit{flexi}-kernel, which reduce metrication artifacts in predictions while at the same time retaining the same hardware efficiency as the box-shaped kernel (details in Appendix I).
\newline
\textbf{Test-Time Augmentation:} To increase robustness of extracted features, we extend a feature-ensembling approach for image classification \cite{ashukha21} to our OW-IS framework by aligning and averaging $(h)^l$ obtained from several different augmentations. In this work, augmentations are rotations of $x_q$ that are restricted to orthogonal increments along each axis (i.e., $\{0^\circ, 90^\circ, 180^\circ, 270^\circ\}$). Rotated variants are embedded, aligned, and averaged with the original $(h)^l$ to obtain a single augmented feature pyramid. To ensure consistency, a fixed augmentation scheme is applied to both queries and memory samples. Notably, combining feature pyramids is much faster than our original TTA approach combining predictions \cite{winwim24}, as a call to $\mathcal{G}$ now has constant time complexity $\mathcal{O}(1)$ instead of linear complexity $\mathcal{O}(a)$ w.r.t. the number of augmentations $a$. This is especially relevant for the alternating interaction phase, where low-latency is required when interacting with $\mathcal{U}$.
\newline
\textbf{Exploration vs. Exploitation in $k$-NN Search:} Let us consider a fixed level $l$ with the total number of query voxels \mbox{$p=\prod_d\Omega^l$}. With \parm, $k$-NN search between $p$ queries and $n$ stored memory patterns is performed independently for all queries using cosine similarity. This corresponds to greedy search with the goal to find the best match for every query voxel. However, greedy search typically leads to the hubness problem \cite{radovanovic10hubness}, where, in high-dimensional $k$-NN search a small subset of patterns ("hubs") frequently appear as NNs to an unusually large number of query voxels. Moreover, there is no ground truth 1:1 mapping between queries and stored patterns in the case of unpaired correspondence matching. Dual-softmax matching \cite{sun21loftr,edstedt24roma} is therefore not applicable in our case.

Instead, we propose regularized cosine matching using a single softmax. By combining soft mutual nearest neighbor filtering \cite{rocco18nn} with parameter-free modern Hopfield retrieval~\cite{ramsauer21hopfield} and the regularization term of the dual-softmax matcher~\cite{sun21loftr}, we introduce the following regularizer to resolve hubness and enable an exploration vs. exploitation trade-off for unpaired correspondence matching:
\begin{equation}
    s_{reg} := \text{softmax}_p \left( \left( \log p \right)^{\beta_{reg}} \cdot s_{cos}(h_q, h_j) \right).
\end{equation}
This regularizer re-distributes matching weights over all $p$ queries (also see Appendix I, Algorithm 1). The commitment hyperparameter $\beta_{reg}$ determines how exclusive a stored memory pattern is to an individual query voxel w.r.t. all other query voxels. Since $k$-NN search is carried out hierarchically and locally across memory feature pyramids, the commitment factor $\left( \log p \right)^{\beta_{reg}}$ concentrates the matching process on spatially local regions of query feature maps.

\section{Experiments}
\label{sec:experiments}
In this section we present our experiments split across three domains: closed-set, open-set, and open-world.

\textbf{Data:} We consider different datasets in our experiments, covering a diverse set of anatomical structures and modalities, as well as non-medical 3D data. Importantly, our dataset choices allow addressing the four categories of known and unknown classes in the OW setting as defined in \cref{method:openworld}. For performance analysis on known classes (KKCs), we use the organ classes liver, kidney, pancreas, and spleen from the \totseg \cite{wasserthal23totseg}, \msd \cite{antonelli22msd}, and \kits\ \cite{heller23kits} datasets, respectively. For performance analysis on unknown classes and domains (KUCs, UKCs and UUCs), we use the \mousemicro\ \cite{rosenhain18} and \antbrain\ \cite{toulkeridou23insect} $\mu$CT datasets, as well as the \brats\ \cite{antonelli22msd} MRI brain tumor dataset. We provide detailed information about used datasets and splits in Appendix II.

\textbf{Virtual User:} For automated evaluation, we employ a virtual user $\mathcal{U}_v$ whose role is mimicking typical behavior of a human user $\mathcal{U}_h$ (see Appendix III for details). By default, $\mathcal{U}_v$ has access to ground truth class indices $c_{gt}$ and segmentation masks $z_{gt}$ for simulating clicks and adjusting interactive hyperparameters $\boldsymbol{\iota}$. For fair comparisons to baseline methods, we let $\mathcal{U}_v$ only perform clicks without any brushing.

\textbf{Metrics:} We use the Dice-Sørensen coefficient \cite{dice45,sorensen48} ($DSC$) for measuring segmentation quality. Interaction with a virtual user $\mathcal{U}_v$ stops when $DSC_{min}=0.95$ is reached within a maximum of $n_{max}=8$ clicks for segmenting any SoI. We define $DSC^*$ as the segmentation quality after $n$ clicks (e.g., $DSC^8$ for $n=8$). The early success rate ($ER$, higher is better) is the proportion of SoIs that require $< n$ clicks before interaction stops, i.e., reaching minimum performance before $n$ clicks. The number of clicks metric ($NoC^*$ \cite{jang19noc}) measures how many clicks were needed to reach a certain performance (e.g., $NoC^{90}$ for $DSC=0.9$). In addition to $NoC^*$, we measure the success rate ($SR$, higher is better), defined as the fraction of successfully segmented SoIs (i.e., those who reached $DSC=*$) relative to the total number of evaluated SoIs. Since $NoC^*$ is exclusively computed on successfully segmented SoIs, reporting it without $SR$ can lead to biased or misleading interpretations. To address known limitations of $NoC^*$ and $DSC^*$ \cite{forte20,moskalenko24}, we propose two novel effort-aware metrics -- rate-penalized $NoC_r^*$ and rate-corrected $DSC_r^*$: 
\begin{align}
    NoC_r^* &= SR \cdot NoC^* + (1-SR) \cdot n_{max}\ ,\\
    DSC_r^* &= ER \cdot DSC_{final} + (1-ER) \cdot DSC^*\ .
\end{align}
$NoC_r^*$ reflects the effort that was spent on fails (i.e., not reaching the respective performance within $n_{max}$ clicks), while $DSC_r^*$ reflects the contribution from early successes. As stated in \cref{method:problem}, our overall objective is maximizing segmentation quality w.r.t. an SoI while minimizing required user guidance. To account for this, we use the cumulative mean performance after consecutive user actions:
\begin{equation}
    DSC_{cr}^n=\frac{1}{n}\sum_{i=1}^n DSC_r^i\ .
\end{equation} 
This metric aligns with recent views that area under the curve metrics are much more relevant for the task of IS \cite{forte20}.

\textbf{Baselines:} As the first to propose OW-IS, respective baseline methods are absent. We are therefore limited to comparing our method against existing OS-IS baselines, as well as performing extensive ablation studies in the OW case. For consistent comparisons with baselines, we integrate \nninter \cite{isensee25nninter}, \sammed \cite{wang25sammed3d}, and \vista\ \cite{he24vista3d} into our framework and apply their default hyperparameters. Although these baseline methods were trained on multiple domains, they lack explicit strategies for continually learning new domains and classes without encountering the catastrophic forgetting problem. Given that \nninter\ was trained on $30.7\times$, $53.6\times$, and $5.6\times$ more data relative to feature extractors used for \meow\ (see \cref{tab:efficiency_comparison}), we consider \nninter\ as an empirical upper bound on segmentation performance.

\textbf{\meow\ Setup:} To demonstrate that our method is backbone-agnostic and compatible with various hourglass-style architectures, we evaluate it using three distinct feature extractors: \suprem\ ($\mathcal{F}_1$) \cite{li24unetsuprem}, \totseg\ ($\mathcal{F}_2$) \cite{wasserthal23totseg}, and the automatic-branch sub-module of \vista\ ($\mathcal{F}_3$) \cite{he24vista3d}. Note that unlike our baseline methods, neither feature extractor is pre-trained explicitly for IS, but for multi-class semantic segmentation. Note that feature extractors requiring weak user labels as inputs (e.g., \nninter) are incompatible with \meow, as they would necessitate repeated feature extraction after every user interaction. For test-time augmentation (TTA), we augment both queries and memory samples by averaging the original feature map $(h)^l$ in addition to two $90^\circ$ rotated variants (rotated around the depth and width axes, respectively). In the presence of $\mathcal{M}$, we store $m=64$ uncompressed (i.e., $L_{sp}=0$) memory samples by default. For 3D data, we set the $k$-NN search reduction factor $\phi=0.25$ in order to be scalable to high resolutions while retaining quasi-constant hardware memory consumption \cite{winwim24}. Unless stated otherwise, we apply all improvements discussed in \cref{method:improving_parmesan}.

\begin{table}[t]
\scriptsize
\centering
\caption{Model size and training data efficiency of different methods. For \meow\, feature extractors are frozen (marked with *).}
\begin{tabular}{lp{1.1cm}p{1cm}p{1cm}p{1cm}}
\hline
  \textbf{Method} & \textbf{Learnable Params} & \textbf{Training Volumes} & \textbf{Training Masks} & \textbf{Memory Samples} \\
  \hline
  \sammed & 100.51 M & 22 K & 143 K & \noneleft\\
  \vista & 217.\underline{96} M & 11.5 K & $>$1.34 M & \noneleft\\
  \nninter & 102.36 M & $\sim$61.3 K & $\sim$681 K & \noneleft\\
  \meow\ ($\mathcal{F}_1\& \mathcal{M}$) & \textbf{*19.07 M} & 2.1 K & $\sim$\textbf{52 K} & $\le128$ \\
  \meow\ ($\mathcal{F}_2\& \mathcal{M}$) & *31.21 M & \textbf{1.2 K} & $<$141 K & $\le128$ \\
  \meow\ ($\mathcal{F}_3\& \mathcal{M}$) & *217.\underline{26} M & 11.5 K & $>$1.34 M & $\le128$ \\
\end{tabular}
\label{tab:efficiency_comparison}
\end{table}

\textbf{Hyperparameters:} To find an optimal set of interactive and non-interactive hyperparameters, we leverage Bayesian optimization \cite{shahriari16bayes}. We use organ KKCs (i.e., liver, pancreas, spleen, and kidney) for evaluating our objective function $f(\textbf{x})$ on a small, independent training split. We maximize $DSC_{cr}^8$ until convergence by optimizing the parameter set $\textbf{x}=\{\tau, \omega, \gamma, \xi^-, \beta_{nn}, \beta_{reg}, \alpha_{fuse}, \sigma_{mp}, \lambda_{mp}, L_{mp}, \beta_{mp}, \eta_{mp}, \text{TTA}\}$ (see Appendix IV for details).

\textbf{Technical Setup:} All methods including baselines are implemented in a unified PyTorch~\cite{paszkepytorch} framework \cite{perez21tio,hunter07,harris20np,markiewicz26,silversmith21cc3d} and evaluated using individual NVIDIA GPUs (models RTX~3090, A40, or A100). Every experimental setup satisfies a memory constraint of $\le 24$~GB VRAM and all timing measurements are benchmarked on a single workstation featuring an RTX~3090, reflecting a consumer-grade hardware setup.

\subsection{Closed-Set Interactive Segmentation}

\textbf{General Improvements:} Improving upon \parm, we introduce several architectural modifications that increase overall performance. To evaluate individual contributions, we conduct extensive ablation studies on the \msd+\kits\ organ CT data and summarize our results in~\cref{tab:ablations_parmesan}. Transitioning from original settings to our proposed settings leads to substantial performance gains. The largest performance gains are obtained from feature compression and new kernels. Replacing simple channel trimming with a linear autoencoder preserves representative feature information (A1, A9, A17). Replacing the $2^{d}$ box-shaped kernel with our improved kernels allows for better correspondence matching (A2, A10, A18). While individual adjustments to similarity weighting, balancing, and regularized matching yield mixed results, synergistic effects (A7, A15, A23) suggest that the interplay between all components is crucial for robust IS.

\begin{table}[t]
\scriptsize
\centering
\caption{Results for ablation studies, demonstrating various improvements when applying modifications to \parm.}
\begin{tabular}{p{0.35cm}llll}
\hline
\textbf{ID} & \textbf{Method} & \textbf{Ablation} & \textbf{Original $\rightarrow$ Ours} & $\mathbf{DSC_{cr}^8\uparrow(ER\uparrow)}$\\
\hline
A0 & $\mathcal{F}_1 \& \mathcal{M}$ & \parm & all original & $0.805\ (0.106)$ \\
A1 & $\mathcal{F}_1 \& \mathcal{M}$ & compression & trim $\rightarrow$ linear AE & $0.851\ (0.234)$ \\
A2 & $\mathcal{F}_1 \& \mathcal{M}$ & kernel & box $\rightarrow$ \textit{cross}+\textit{flexi} & $0.859\ (0.170)$ \\
A3 & $\mathcal{F}_1 \& \mathcal{M}$ & similarity & $\omega=1 \rightarrow 0.90$ & $0.814\ (0.096)$ \\
A4 & $\mathcal{F}_1 \& \mathcal{M}$ & balancing & $\gamma=0 \rightarrow -0.99$ & $0.808\ (0.106)$ \\
A5 & $\mathcal{F}_1 \& \mathcal{M}$ & matcher & cos $\rightarrow$ reg. cos & $0.813\ (0.125)$ \\
A6 & $\mathcal{F}_1 \& \mathcal{M}$ & HMP & $L_{mp}=1$$\rightarrow 2$ & $0.823\ (0.115)$ \\
A7 & $\mathcal{F}_1 \& \mathcal{M}$ & \meow & all ours & $\mathbf{0.892\ (0.541)}$ \\
\hline
A8 & $\mathcal{F}_2 \& \mathcal{M}$ & \parm & all original & $0.660\ (0.0)$ \\
A9 & $\mathcal{F}_2 \& \mathcal{M}$ & compression & trim $\rightarrow$ linear AE & $0.784\ (0.106)$ \\
A10 & $\mathcal{F}_2 \& \mathcal{M}$ & kernel & box $\rightarrow$ \textit{cross}+\textit{flexi} & $0.752\ (0.019)$ \\
A11 & $\mathcal{F}_2 \& \mathcal{M}$ & similarity & $\omega=1 \rightarrow 0.50$ & $0.713\ (0.0)$ \\
A12 & $\mathcal{F}_2 \& \mathcal{M}$ & balancing & $\gamma=0 \rightarrow -0.99$ & $0.658\ (0.0)$ \\
A13 & $\mathcal{F}_2 \& \mathcal{M}$ & matcher & cos $\rightarrow$ reg. cos & $0.653\ (0.0)$ \\
A14 & $\mathcal{F}_2 \& \mathcal{M}$ & HMP & $L_{mp}=1$$\rightarrow 2$ & $0.685\ (0.0)$ \\
A15 & $\mathcal{F}_2 \& \mathcal{M}$ & \meow & all ours & $\mathbf{0.836\ (0.199)}$ \\
\hline
A16 & $\mathcal{F}_3 \& \mathcal{M}$ & \parm & all original & $0.662\ (0.010)$ \\
A17 & $\mathcal{F}_3 \& \mathcal{M}$ & compression & trim $\rightarrow$ linear AE & $0.737\ (0.019)$ \\
A18 & $\mathcal{F}_3 \& \mathcal{M}$ & kernel & box $\rightarrow$ \textit{cross}+\textit{flexi} & $0.723\ (0.019)$ \\
A19 & $\mathcal{F}_3 \& \mathcal{M}$ & similarity & $\omega=1 \rightarrow 0.65$ & $0.697\ (0.0)$ \\
A20 & $\mathcal{F}_3 \& \mathcal{M}$ & balancing & $\gamma=0 \rightarrow -0.55$ & $0.653\ (0.010)$ \\
A21 & $\mathcal{F}_3 \& \mathcal{M}$ & matcher & cos $\rightarrow$ reg. cos & $0.678\ (0.010)$ \\
A22 & $\mathcal{F}_3 \& \mathcal{M}$ & HMP & $L_{mp}=1$$\rightarrow 2$ & $0.696\ (0.010)$ \\
A23 & $\mathcal{F}_3 \& \mathcal{M}$ & \meow & all ours & $\mathbf{0.789\ (0.048)}$ \\
\end{tabular}
\label{tab:ablations_parmesan}
\end{table}

\begin{table}[!ht]
\scriptsize
\centering
\caption{Results for ablation studies (top block: $DSC_{cr}^8$ bottom block: $DSC_r^8$), demonstrating the effectiveness of HMP under strong (mass-conserving) perturbations of 3D base predictions $y_b$ preceding HMP.}
\begin{tabular}{l|llllll}
$\mathbf{L_{mp}}$ & \textbf{Unperturbed} & \textbf{Blur} & \textbf{Noise} & \textbf{Translate} & \textbf{Rotate} & \textbf{Shear} \\
\hline
0 & $0.814$ & $0.655$ & $0.403$ & $0.531$ & $0.364$ & $0.492$ \\
1 & $0.830$ & $0.712$ & $0.828$ & $0.578$ & $0.394$ & $0.563$ \\
2 & $\mathbf{0.840}$ & $0.731$ & $\mathbf{0.829}$ & $0.644$ & $0.448$ & $0.639$ \\
3 & $\mathbf{0.840}$ & $\mathbf{0.746}$ & $0.823$ & $\mathbf{0.710}$ & $\mathbf{0.521}$ & $\mathbf{0.691}$ \\
\hline
0 & $0.833$ & $0.707$ & $0.487$ & $0.554$ & $0.361$ & $0.499$ \\
1 & $0.856$ & $0.703$ & $0.851$ & $0.612$ & $0.390$ & $0.571$ \\
2 & $0.868$ & $0.712$ & $\mathbf{0.855}$ & $0.686$ & $0.444$ & $0.647$ \\
3 & $\mathbf{0.872}$ & $\mathbf{0.732}$ & $0.851$ & $\mathbf{0.749}$ & $\mathbf{0.515}$ & $\mathbf{0.704}$ \\
\end{tabular}
\label{tab:ablations_hmp}
\end{table}

\begin{figure}[!ht]
	\centering
	\includegraphics[width=1.\linewidth]{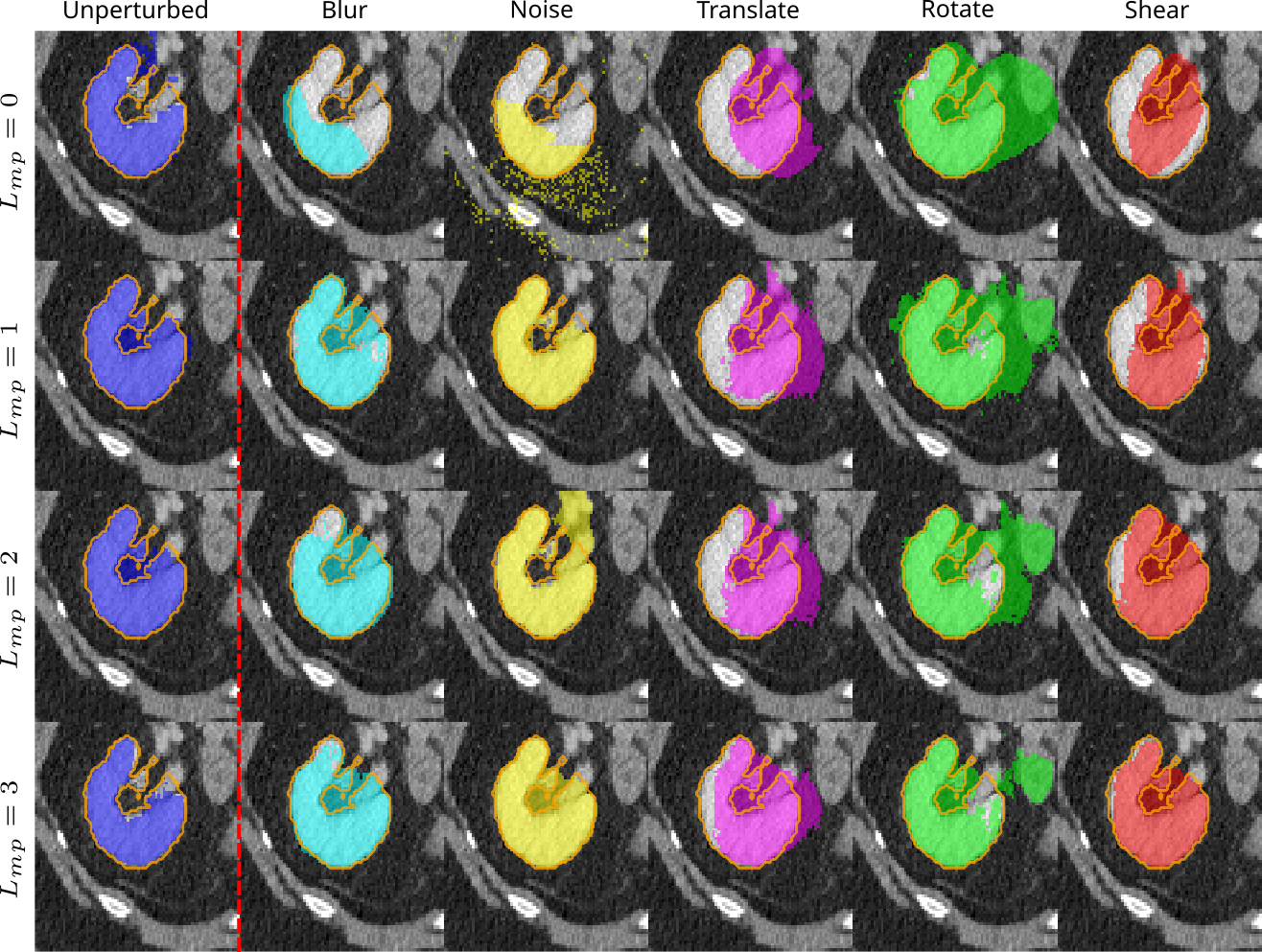}
	\caption{Qualitative results demonstrating the effectiveness of hierarchical message passing (HMP) under strong (mass-conserving) perturbations of 3D base predictions $y_b$ preceding HMP. Ground truth is indicated in contours.}
	\label{fig:ablations_hmp}
\end{figure}

To demonstrate the effectiveness of hierarchical message passing (HMP), we conduct ablation studies with degraded base predictions $y_b$ preceding HMP using $\mathcal{F}_1$ without $\mathcal{M}$. We use perturbed base predictions to isolate the contribution of HMP, test robustness under realistic conditions, and highlight the benefit of hierarchical vs. single-level MP. Our results are presented in \cref{tab:ablations_hmp} and \cref{fig:ablations_hmp}. While the original \parm\ approach used $L_{mp}=1$ to correct minor artifacts, increasing $L_{mp}$ allows to correct strong perturbations of base predictions. This iterative correction is consistent with the behavior of message-passing graph neural networks, which can be viewed as performing a form of local fixed-point inference on the voxel graph \cite{giovanni22graph,zhou03transduction}.

\vspace{\baselineskip}
\textbf{Interactive Segmentation:} We evaluate closed-set performance of all methods on organ CT classes (KKCs) and report results in \cref{tab:KKCs}. We achieve state of the art performance on the \msd+\kits\ organ segmentation task (B1.3). \meow\ otherwise performs similar to \sammed\ and \vista\ baselines with feature extractors pre-trained on substantially less data (see \cref{tab:efficiency_comparison}). Using $\mathcal{M}$ increases performance in all setups while reducing the required $NoC$. This is due to memory predictions providing high-quality segmentations before alternating user interaction, meaning that $\mathcal{U}$ only needs to provide corrective guidance during the interaction phase (also see $DSC_{cr}^1$ column in \cref{tab:KKCs}). \meow\ is robust w.r.t. the choice of $\mathcal{F}$, with $\mathcal{F}_1$ (\suprem) yielding best overall results. \meow\ is almost on par with \vista\ while not using its decoder (B0.1, B0.7, B1.1, B1.7).

\nninter\ achieves the best overall performance across datasets. However, our framework offers distinct structural advantages for real-world deployment. First, \meow\ is an OW-IS method, whereas \nninter\ is designed as an OS-IS method without dedicated OW-learning strategies. Second, our system is inherently more flexible, featuring a modular design that allows for the seamless replacement of backbones and memory modules without re-training. Third, our approach is designed for fast learning and stable knowledge retention. Moreover, \nninter\ was trained on a vastly larger training corpus relative to $\mathcal{F}_1$, $\mathcal{F}_2$, and $\mathcal{F}_3$ (see \cref{tab:efficiency_comparison}), underlining \meow's data-efficiency. These backbones were further optimized for multi-class segmentation rather than IS. This introduces an additional transfer learning task that we successfully resolve, thereby enabling the system's superior adaptability and OW-learning capabilities.

\begin{table}[t]
\scriptsize
\centering
\caption{Closed-set IS performance results for $\text{KKCs}$ on medical datasets (top: \totsegdata, bottom: \msd+\kits) for organ classes liver, kidney, pancreas, and spleen.}
\begin{tabular}{p{0.35cm}p{1.9cm}p{0.9cm}p{0.9cm}p{0.9cm}p{0.9cm}}
\hline
\textbf{ID} & \textbf{Method} & $\mathbf{NoC_r^{90}\downarrow}$ & $\mathbf{DSC_r^8\uparrow}$ & $\mathbf{DSC_{cr}^1\uparrow}$ & $\mathbf{DSC_{cr}^8\uparrow}$\\
\hline
B0.0 &  \nninter & $\mathbf{2.518}$ & $\mathbf{0.940}$ & $\mathbf{0.854}$ & $\mathbf{0.920}$ \\
B0.1 &  \vista & $5.296$ & $0.844$ & $0.702$ & $0.797$ \\
B0.2 &  \sammed & $4.100$ & $0.847$ & $0.805$ & $0.842$ \\
B0.3 &  \meow\ ($\mathcal{F}_1 \& \mathcal{M}$) & $4.184$ & $0.858$ & $0.792$ & $0.831$ \\
B0.4 &  \meow\ ($\mathcal{F}_1$) & $5.321$ & $0.795$ & $0.637$ & $0.740$ \\
B0.5 &  \meow\ ($\mathcal{F}_2 \& \mathcal{M}$) & $4.529$ & $0.838$ & $0.767$ & $0.817$ \\
B0.6 &  \meow\ ($\mathcal{F}_2$) & $5.870$ & $0.779$ & $0.593$ & $0.733$ \\
B0.7 &  \meow\ ($\mathcal{F}_3 \& \mathcal{M}$) & $6.322$ & $0.786$ & $0.676$ & $0.751$ \\
B0.8 &  \meow\ ($\mathcal{F}_3$) & $7.640$ & $0.692$ & $0.469$ & $0.628$ \\
\hline
B1.0 &  \nninter & $\mathbf{3.106}$ & $\mathbf{0.918}$ & $0.783$ & $0.886$ \\
B1.1 &  \vista & $5.436$ & $0.834$ & $0.747$ & $0.811$ \\
B1.2 &  \sammed & $3.585$ & $0.856$ & $0.772$ & $0.847$ \\
B1.3 &  \meow\ ($\mathcal{F}_1 \& \mathcal{M}$) & $3.316$ & $0.903$ & $\mathbf{0.868}$ & $\mathbf{0.892}$ \\
B1.4 &  \meow\ ($\mathcal{F}_1$) & $3.928$ & $0.873$ & $0.781$ & $0.841$ \\
B1.5 &  \meow\ ($\mathcal{F}_2 \& \mathcal{M}$) & $4.021$ & $0.857$ & $0.787$ & $0.836$ \\
B1.6 &  \meow\ ($\mathcal{F}_2$) & $5.272$ & $0.811$ & $0.634$ & $0.762$ \\
B1.7 &  \meow\ ($\mathcal{F}_3 \& \mathcal{M}$) & $6.028$ & $0.816$ & $0.738$ & $0.790$ \\
B1.8 &  \meow\ ($\mathcal{F}_3$) & $7.0$ & $0.736$ & $0.502$ & $0.673$ \\
\end{tabular}
\label{tab:KKCs}
\end{table}

\subsection{Open-Set Interactive Segmentation}

We now investigate open-set (OS) capabilities of our method, i.e., segmenting unknown classes in the context of $\text{UKCs}$ and $\text{UUCs}$. We select \mousemicro\ \cite{rosenhain18} and \antbrain\ \cite{toulkeridou23insect} datasets for quantitative evaluation, since none of our baselines, nor feature extractors were pre-trained on them. Although \nninter\ was trained on different micro-CT data, we still consider our selected sets as unknown to this baseline. Performance results are summarized in \cref{tab:os_is}. Note that for \meow, $\mathcal{M}$ is not incorporated in this setting, since the pure OS scenario provides no samples to store. The working memory $M$ is therefore filled exclusively with the query itself for interpreting user guidance.

\begin{table}[t]
\scriptsize
\centering
\caption{Open-set IS performance results for $\text{UKCs}$ for unknown CT datasets (top: \mousemicro, bottom: \antbrain). We use all organ classes for \mousemicro\ and the single brain class for \antbrain. Note that we do not use $\mathcal{M}$ for \meow.}
\begin{tabular}{p{0.35cm}p{1.5cm}llll}
\hline
\textbf{ID} & \textbf{Method} & $\mathbf{NoC_r^{90}\downarrow}$ & $\mathbf{DSC_r^8\uparrow}$ & $\mathbf{DSC_{cr}^1\uparrow}$ & $\mathbf{DSC_{cr}^8\uparrow}$\\
\hline
B2.0 & \nninter & $\mathbf{4.880}$ & $\mathbf{0.894}$ & $\mathbf{0.650}$ & $\mathbf{0.826}$ \\
B2.1 & \vista & $7.990$ & $0.537$ & $0.166$ & $0.422$ \\
B2.2 & \sammed & $7.698$ & $0.673$ & $0.485$ & $0.651$ \\
B2.3 & \meow\ ($\mathcal{F}_1$) & $7.760$ & $0.747$ & $0.495$ & $0.668$ \\
B2.4 & \meow\ ($\mathcal{F}_2$) & $7.990$ & $0.694$ & $0.340$ & $0.583$ \\
B2.5 & \meow\ ($\mathcal{F}_3$) & $7.990$ & $0.710$ & $0.474$ & $0.622$ \\
\hline
B3.0 &  \nninter & $\mathbf{4.280}$ & $\mathbf{0.940}$ & $\mathbf{0.629}$ & $\mathbf{0.859}$ \\
B3.1 &  \vista & $7.880$ & $0.664$ & $0.209$ & $0.503$ \\
B3.2 &  \sammed & $8.0$ & $0.708$ & $0.522$ & $0.688$ \\
B3.3 &  \meow\ ($\mathcal{F}_1$) & $7.800$ & $0.790$ & $0.537$ & $0.715$ \\
B3.4 &  \meow\ ($\mathcal{F}_2$) & $7.960$ & $0.747$ & $0.553$ & $0.684$ \\
B3.5 &  \meow\ ($\mathcal{F}_3$) & $8.0$ & $0.719$ & $0.478$ & $0.671$ \\
\end{tabular}
\label{tab:os_is}
\end{table}

In the OS scenario, our framework systematically outperforms the \vista\ and \sammed\ baselines. Notably, using \meow\ with a \vista\ sub-module for feature extraction ($\mathcal{F}_3$) yields a substantial performance gain over the full \vista\ method (B2.1, B2.5, B3.1, B3.5). These results demonstrate that the learned interactive branch in \vista\ can be effectively replaced by our parameter-free transduction approach, simultaneously reducing learnable parameters, enabling dynamic hyperparameters, and enhancing OS-IS performance.

\nninter\ performs best among all evaluated methods for the reasons discussed above. Crucially, \meow\ is not being leveraged to its full architectural potential in this specific comparison, partially accounting for this performance gap and identifying the integration of $\mathcal{M}$ as a critical mechanism.

\begin{figure}[t]
	\centering
	\includegraphics[width=1.\linewidth]{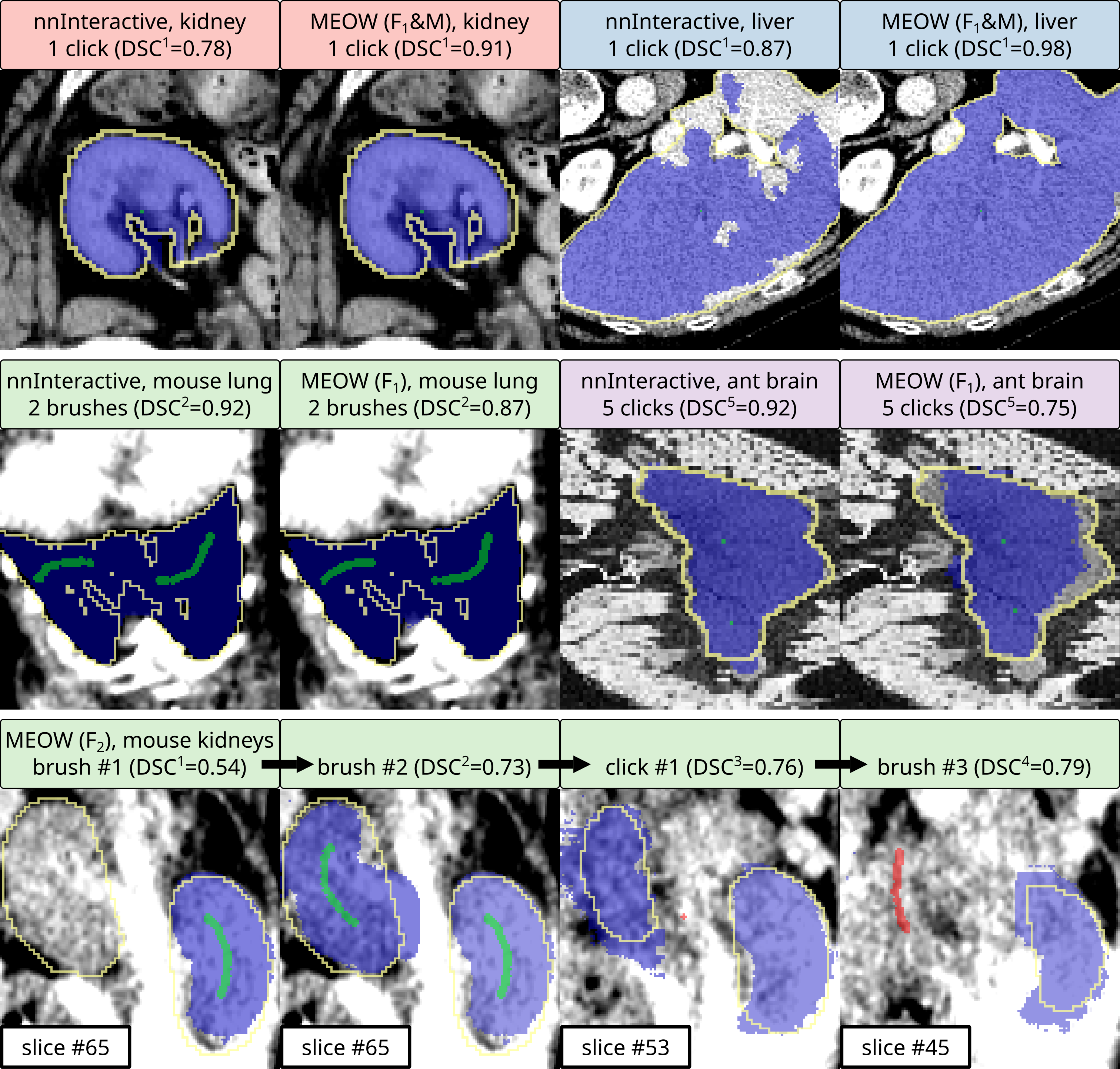}
	\caption{Qualitative demonstrations of user interactions (blue: model prediction, green/red: weak positive/negative labels $z^\pm$, contours: ground truth). With \meow, clicks and brushes can be mixed flexibly across slices and axes.}
	\label{fig:interactionplot}
\end{figure}

\begin{figure}[t]
	\centering
    \includegraphics[width=1.\linewidth]{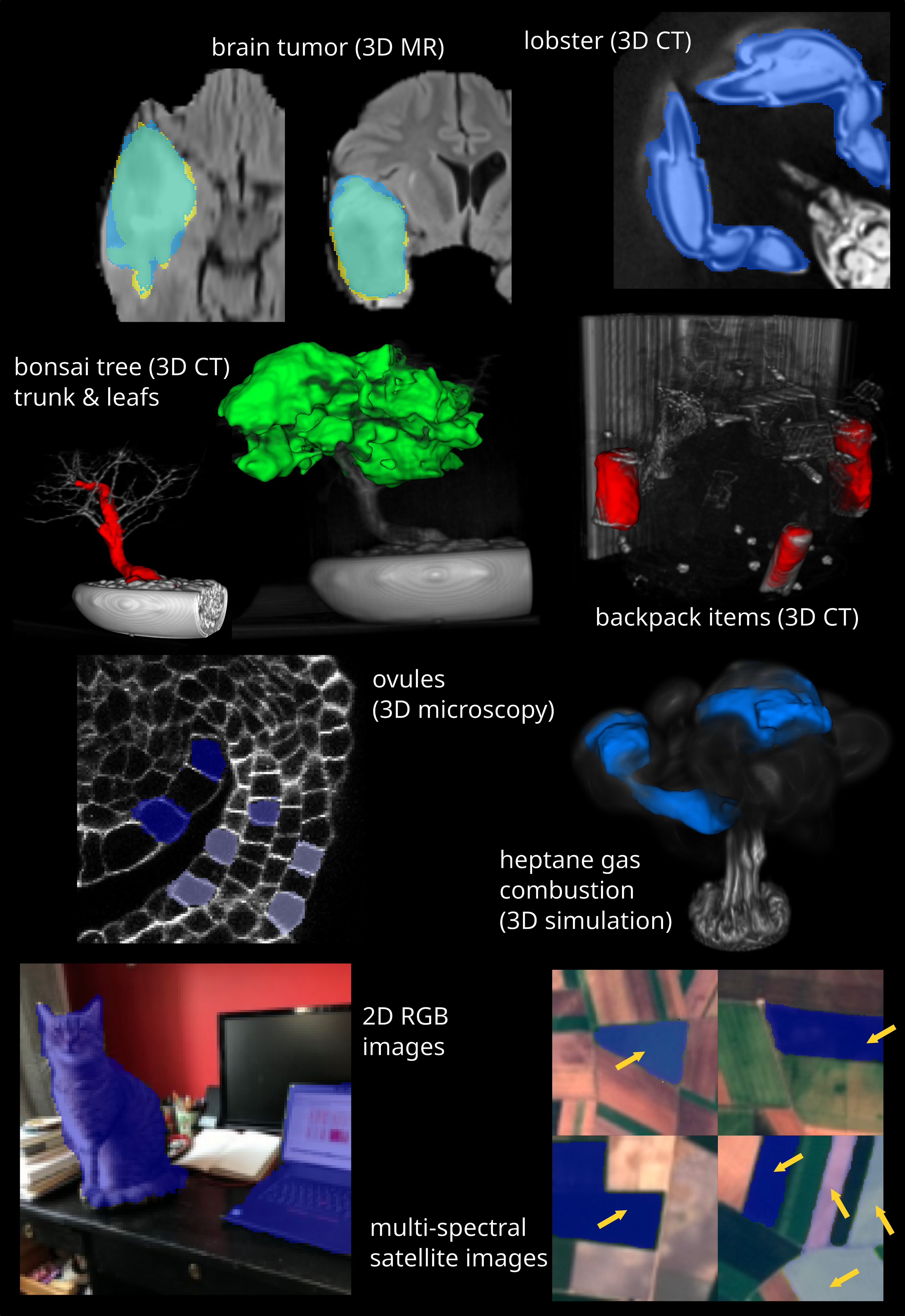}
	\caption{Qualitative demonstrations of open-domain IS capabilities of \meow\ (predictions in colors). We use \suprem\ pre-trained on human abdomen CT data (i.e., $\mathcal{F}_1$ without $\mathcal{M}$) to segment various unknown objects under extreme domain shifts, as well as different data modalities  \cite{antonelli22msd,scivisdata,zhou25_dataovules,helber18eurosat}. Individual results were obtained in $\approx\!1$ minute of human user interaction. For the brain tumor example, prediction and ground truth are visualized in blue and yellow, respectively ($DSC^4=0.823$).}
	\label{fig:collage}
\end{figure}

We perform qualitative tests of \meow\ w.r.t. user guidance and visualize results in \cref{fig:interactionplot}. We demonstrate IS on KKCs and UKCs using diverse user actions. Moreover, we demonstrate IS under increasingly extreme domain shifts and visualize qualitative results in \cref{fig:collage}.

\meow\ is capable of instance classification by leveraging stored samples with known metadata (i.e., KKCs / UKCs). Given 4 organ classes liver, kidney, pancreas, and spleen, we find instance classification accuracies of $0.909$ (B0.3), $0.966$ (B0.5), $0.923$ (B0.7), $0.993$ (B1.3), $0.999$ (B1.5), and $0.986$ (B1.7) (see \cref{tab:KKCs}).

\textbf{Real-Time Capabilities:} IS methods should be designed to interact with $\mathcal{U}$ in real-time. We therefore report inference speed $t_{inf}$ and learning speed $t_{learn}$ (if applicable) in \cref{tab:speed}. We consider any method with $t_{inf}<1$ seconds as real-time IS and any method with $t_{learn}<1$ seconds as real-time learning. \meow\ meets both of these conditions. Notably, our approach can learn a sample as fast as $t_{learn}=7.61 \pm 0.18$ milliseconds through memory consolidation, substantially outperforming standard fine-tuning strategies that typically require several minutes to hours \cite{phuntsho25limitations}.

\begin{table}[t]
\scriptsize
\centering
\caption{Mean and standard deviation of inference speed $t_{inf}$ (per model prediction) and learning speed $t_{learn}$ (per sample) at a resolution of $128^3$ voxels using an RTX~3090 GPU. \meow\ is real-time capable w.r.t. inference and learning using frozen feature extractors.}
\begin{tabular}{p{0.35cm}llll}
\hline
\textbf{ID} & \textbf{Method} & $\mathbf{t_{inf}}$ \textbf{[ms]} & $\mathbf{t_{learn}}$ \textbf{[ms]} \\
\hline
B0.0 & \nninter & $289.750 \pm 4.610$ & \none \\
B0.1 & \vista & $267.430 \pm 6.647$ & \none \\
B0.2 & \sammed & $\mathbf{8.980 \pm 0.099}$ & \none \\
B0.3 & \meow ($\mathcal{F}_1 \& \mathcal{M}$) & $279.470 \pm 12.488$ & $9.62 \pm 0.11$ \\
B0.4 & \meow ($\mathcal{F}_1$) & $519.360 \pm 15.613$ & \none \\
B0.5 & \meow ($\mathcal{F}_2 \& \mathcal{M}$) & $501.315 \pm 13.018$ & $\mathbf{7.61 \pm 0.18}$ \\
B0.6 & \meow ($\mathcal{F}_2$) & $517.205 \pm 1.846$ & \none \\
B0.7 & \meow ($\mathcal{F}_3 \& \mathcal{M}$) & $594.450 \pm 0.438$ & $57.75 \pm 0.71$ \\
B0.8 & \meow ($\mathcal{F}_3$) & $625.935 \pm 1.011$ & \none \\
\end{tabular}
\label{tab:speed}
\end{table}

\subsection{Open-World Interactive Segmentation}
We now investigate open-world (OW) capabilities of our method, i.e., continually learning and improving by leveraging model predictions from unknown classes. Recall that OW recognition requires having open-set (OS) capabilities, i.e., being able to make predictions for unknown classes in the first place. OW-learning is achieved by first predicting pseudo-labels $z_p$ for incoming queries, followed by memory consolidation in $\mathcal{M}$. As \meow\ represents the first method to address the OW-IS problem, comparisons to baseline methods are limited to the OS-IS case, which translates to using \meow\ without $\mathcal{M}$ ($m=0$).

During evaluation, we turn off any learning and evaluate on a fixed validation split. During training in an OW scenario, query images along with their final segmentation results are being learned one at a time. Since \meow\ applies transductive reasoning from $\mathcal{M}$, learned samples are instantly leveraged for processing any subsequent query. We report mean and standard deviation of performance measurements over two independent training folds (more details in Appendix II). Unless stated otherwise, our default setup for OW-IS experiments is $L_{sp}=1$ (sparse), $m=128$, $z_p$ in $\mathcal{M}$, treating samples as $\text{UKCs}$ (specifically $\mathcal{U}$ having access to class indices), and using pre-trained $\mathcal{F}$.

\begin{figure*}[t]
	\centering
	\includegraphics[width=1.\linewidth]{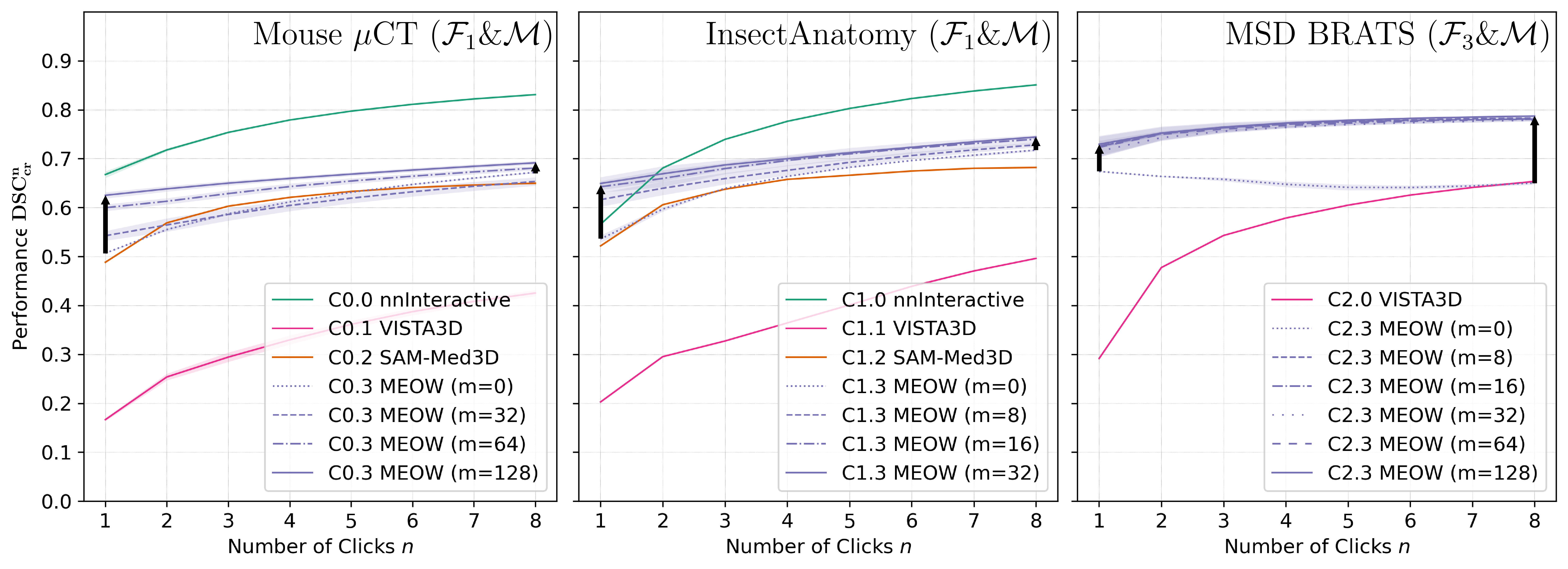}
	\caption{Demonstration of OW-learning for IS via memory consolidation. \meow\ is continually trained by filling $\mathcal{M}$ with self-generated pseudo-labels, which are then used for making future predictions via transductive reasoning. OW-learning typically leads to increased performance (black arrows), even with small memory modules.}
	\label{fig:ow_main}
\end{figure*}

We present OW-IS experiment results in \cref{fig:ow_main} and \cref{tab:ow_main}. OW-learning generally improves segmentation performance for all feature extractors $\mathcal{F}$. Increasing memory size $m$ consistently improves results, though the benefit of OW-learning takes effect already at very small $m$. For example, we observe relative performance gains of up to $DSC_{cr}^1=0.118$ and $DSC_{cr}^8=0.133$ with as few as $m=8$ stored samples (C1.4, C2.3), underlining \meow's data-efficiency in terms of OW-learning. Because performance usually converges later on, earlier clicks typically benefit more from OW-learning than later clicks. Since memory states are fully reversible, identical results would be obtained in the case of unlearning, i.e., removing samples from $\mathcal{M}$.

\meow\ outperforms \vista\ by a large margin in all OW-IS experiments, even without $\mathcal{M}$ (C0.1, C0.3-5, C1.1, C1.3-5). \meow\ also outperforms \sammed\ already without $\mathcal{M}$ on \mousemicro\ (C0.2, C0.3) and \antbrain\ (C1.2, C1.3). \meow\ generally underperforms \nninter, except for $DSC_{cr}^1$ on \antbrain\ when using $\mathcal{F}_1$ or $\mathcal{F}_2$ with as few as $m=8$ memory samples (C1.0, C1.3, C1.4). This result indicates that \meow\ is highly efficient in terms of required user guidance. 

\begin{table}[t]
  \centering
    \scriptsize
    \centering
    \caption{OW-IS results for OW-learning capabilities on \mousemicro\ data (top) and \antbrain\ data (bottom) with different number $m$ of pseudo-labeled memory samples ($\text{UKCs}$).}
    \begin{tabular}{p{0.35cm}llll}
    \hline
    \textbf{ID} & \textbf{Method} & \textbf{m (\#SoIs)} & $\mathbf{DSC_{cr}^1\uparrow}$ & $\mathbf{DSC_{cr}^8\uparrow}$\\
    \hline
     C0.0 & \nninter & \none & $\mathbf{0.667 \pm 0.007}$ & $\mathbf{0.831 \pm 0.0}$ \\
    C0.1 & \vista & \none & $0.167 \pm 0.003$ & $0.425 \pm 0.006$ \\
    C0.2 & \sammed & \none & $0.488 \pm 0.0$ & $0.650 \pm 0.001$ \\
    C0.3 & \meow\ ($\mathcal{F}_1 \& \mathcal{M}$) & $0$ & $0.507 \pm 0.001$ & $0.672 \pm 0.001$ \\
     &  & $\rightarrow 32$ & $0.542 \pm 0.010$ & $0.653 \pm 0.009$ \\
     &  & $\rightarrow 64$ & $0.600 \pm 0.007$ & $0.681 \pm 0.005$ \\
     &  & $\rightarrow 128$ & $0.625 \pm 0.006$ & $0.691 \pm 0.004$ \\
    C0.4 & \meow\ ($\mathcal{F}_2 \& \mathcal{M}$) & $0$ & $0.336 \pm 0.012$ & $0.576 \pm 0.0$ \\
     &  & $\rightarrow 32$ & $0.451 \pm 0.011$ & $0.528 \pm 0.004$ \\
     &  & $\rightarrow 64$ & $0.500 \pm 0.007$ & $0.563 \pm 0.007$ \\
     &  & $\rightarrow 128$ & $0.515 \pm 0.005$ & $0.574 \pm 0.001$ \\
    C0.5 & \meow\ ($\mathcal{F}_3 \& \mathcal{M}$) & $0$ & $0.468 \pm 0.003$ & $0.613 \pm 0.002$ \\
     &  & $\rightarrow 32$ & $0.361 \pm 0.004$ & $0.559 \pm 0.003$ \\
     &  & $\rightarrow 64$ & $0.411 \pm 0.002$ & $0.588 \pm 0.001$ \\
     &  & $\rightarrow 128$ & $0.440 \pm 0.003$ & $0.608 \pm 0.007$ \\
    \hline
    \hline
    C1.0 & \nninter & \none & $0.566 \pm 0.0$ & $\mathbf{0.851 \pm 0.0}$ \\
    C1.1 & \vista & \none & $0.203 \pm 0.0$ & $0.496 \pm 0.0$ \\
    C1.2 & \sammed & \none & $0.522 \pm 0.0$ & $0.682 \pm 0.0$ \\
    C1.3 & \meow\ ($\mathcal{F}_1 \& \mathcal{M}$) & $0$ & $0.536 \pm 0.008$ & $0.717 \pm 0.002$ \\
     &  & $\rightarrow 8$ & $0.616 \pm 0.014$ & $0.728 \pm 0.005$ \\
     &  & $\rightarrow 16$ & $0.643 \pm 0.012$ & $0.740 \pm 0.009$ \\
     &  & $\rightarrow 32$ & $0.649 \pm 0.013$ & $0.744 \pm 0.002$ \\
     C1.4 & \meow\ ($\mathcal{F}_2 \& \mathcal{M}$) & $0$ & $0.521 \pm 0.014$ & $0.683 \pm 0.004$ \\
     &  & $\rightarrow 8$ & $0.639 \pm 0.016$ & $0.703 \pm 0.010$ \\
     &  & $\rightarrow 16$ & $0.650 \pm 0.009$ & $0.695 \pm 0.016$ \\
     &  & $\rightarrow 32$ & $\mathbf{0.675 \pm 0.010}$ & $0.712 \pm 0.012$ \\
     C1.5 & \meow\ ($\mathcal{F}_3 \& \mathcal{M}$) & $0$ & $0.489 \pm 0.001$ & $0.673 \pm 0.005$ \\
     &  & $\rightarrow 8$ & $0.488 \pm 0.145$ & $0.620 \pm 0.097$ \\
     &  & $\rightarrow 16$ & $0.515 \pm 0.128$ & $0.632 \pm 0.092$ \\
     &  & $\rightarrow 32$ & $0.512 \pm 0.093$ & $0.628 \pm 0.076$ \\
    \end{tabular}
    \label{tab:ow_main}
\end{table}

\begin{table}[t]
    \scriptsize
    \centering
    \caption{OW-IS results for ablation studies on \mousemicro\ data (top) and \antbrain\ data (bottom). We analyze using ground truth labels, using untrained feature extractors (marked with $r$), and using no class metadata.}
    \begin{tabular}{p{0.35cm}lll}
    \hline
    \textbf{ID} & \textbf{Ablation} & $\mathbf{DSC_{cr}^1\uparrow}$ & $\mathbf{DSC_{cr}^8\uparrow}$\\
    \hline
    C0.3 & \none & $\mathbf{0.625 \pm 0.006}$ & $\mathbf{0.691 \pm 0.004}$ \\
    C0.4 & \none & $0.515 \pm 0.005$ & $0.574 \pm 0.001$ \\
    C0.5 & \none & $0.440 \pm 0.003$ & $0.608 \pm 0.007$ \\
    \hline
    C0.6 & $z_p\rightarrow z_{gt}$ in $\mathcal{M}$ ($\mathcal{F}_1$) & $\mathbf{0.682 \pm 0.020}$ & $\mathbf{0.750 \pm 0.009}$ \\
    C0.7 & $z_p\rightarrow z_{gt}$ in $\mathcal{M}$ ($\mathcal{F}_2$) & $0.671 \pm 0.009$ & $0.718 \pm 0.005$ \\
    C0.8 & $z_p\rightarrow z_{gt}$ in $\mathcal{M}$ ($\mathcal{F}_3$) & $0.550 \pm 0.003$ & $0.702 \pm 0.001$ \\
    \hline
    C0.9 & $\mathcal{F}_1^r$ (untrained) & $\mathbf{0.473 \pm 0.005}$ & $\mathbf{0.579 \pm 0.001}$ \\
    C0.10 & $\mathcal{F}_2^r$ (untrained) & $0.392 \pm 0.001$ & $0.530 \pm 0.006$ \\
    C0.11 & $\mathcal{F}_3^r$ (untrained) & $0.459 \pm 0.008$ & $0.530 \pm 0.006$ \\
    \hline
    C0.12 & $\text{UKCs}\rightarrow \text{UUCs}$ ($\mathcal{F}_1$) & $\mathbf{0.595 \pm 0.007}$ & $\mathbf{0.676 \pm 0.0}$ \\
    C0.13 & $\text{UKCs}\rightarrow \text{UUCs}$ ($\mathcal{F}_2$) & $0.471 \pm 0.007$ & $0.567 \pm 0.004$ \\
    C0.14 & $\text{UKCs}\rightarrow \text{UUCs}$ ($\mathcal{F}_3$) & $0.464 \pm 0.011$ & $0.587 \pm 0.003$ \\
    \hline
    \hline
    C1.3 & \none & $0.649 \pm 0.013$ & $\mathbf{0.744 \pm 0.002}$ \\
    C1.4 & \none & $\mathbf{0.675 \pm 0.010}$ & $0.712 \pm 0.012$ \\
    C1.5 & \none & $0.512 \pm 0.093$ & $0.628 \pm 0.076$ \\
    \hline
    C1.6 & $z_p\rightarrow z_{gt}$ in $\mathcal{M}$ ($\mathcal{F}_1$) & $0.651 \pm 0.010$ & $\mathbf{0.749 \pm 0.007}$ \\
    C1.7 & $z_p\rightarrow z_{gt}$ in $\mathcal{M}$ ($\mathcal{F}_2$) & $\mathbf{0.693 \pm 0.003}$ & $0.729 \pm 0.002$ \\
    C1.8 & $z_p\rightarrow z_{gt}$ in $\mathcal{M}$ ($\mathcal{F}_3$) & $0.568 \pm 0.010$ & $0.675 \pm 0.003$ \\
    \hline
    C1.9 & $\mathcal{F}_1^r$ (untrained) & $\mathbf{0.612 \pm 0.002}$ & $\mathbf{0.721 \pm 0.008}$ \\
    C1.10 & $\mathcal{F}_2^r$ (untrained) & $0.448 \pm 0.035$ & $0.546 \pm 0.029$ \\
    C1.11 & $\mathcal{F}_3^r$ (untrained) & $0.552 \pm 0.011$ & $0.655 \pm 0.018$ \\
    \hline
    C1.12 & $\text{UKCs}\rightarrow \text{UUCs}$ ($\mathcal{F}_1$) & $0.642 \pm 0.007$ & $\mathbf{0.747 \pm 0.001}$ \\
    C1.13 & $\text{UKCs}\rightarrow \text{UUCs}$ ($\mathcal{F}_2$) & $\mathbf{0.674 \pm 0.004}$ & $0.728 \pm 0.004$ \\
    C1.14 & $\text{UKCs}\rightarrow \text{UUCs}$ ($\mathcal{F}_3$) & $0.502 \pm 0.073$ & $0.653 \pm 0.039$ \\
    \end{tabular}
    \label{tab:ow_ablations}
\end{table}

\textbf{Ablation Studies:} We further investigate OW-IS capabilities of \meow\ through comprehensive ablation studies (see \cref{tab:ow_ablations}) in order to address aspects regarding:
\begin{itemize}
    \item using $z_{gt}$ instead of $z_p$ in $\mathcal{M}$ to study the performance gain from ground truth labels,
    \item treating samples as $\text{UUCs}$ without any class-information (i.e., relying on novelty detection),
    \item using untrained backbones $\mathcal{F}^r$ with random weights to investigate effectiveness of large-scale pre-training.
\end{itemize}

Using ground truth labels $z_{gt}$ instead of self-generated pseudo-labels $z_p$ tends to slightly increase performance due to degraded pseudo-label quality (C*.6-8). Segmentation performance also remains stable when no metadata regarding an SoI is available (C*.12-14).

We demonstrate that \meow\ is highly feature-robust. Notably, when used in combination with randomized, untrained $\mathcal{F}^r$ (C*.9-11), our proposed framework still outperforms \vista\ across datasets and feature extractors, while \sammed\ is being outperformed in one setup (C1.2 vs. C1.9). Through our hierarchical correspondence matching approach, both our similarity-based IS and transductive reasoning via $\mathcal{M}$ successfully exploits structural priors that are present in image data. We argue that \meow\ can therefore even be applied in data-scarce domains without necessitating any feature extractor pre-training. This finding is supported by related work on random projections \cite{charikar02random,rahimi07random,attaiki23}, reservoir computing \cite{tanaka19reservoir}, distributed representations \cite{plate03repr,kanerva09}, and associative memory \cite{hopfield82,ramsauer21hopfield}.

\textbf{Open-Domain Learning:} OW recognition is not limited to learning new classes in a fixed domain, but also extends to learning within entirely new domains. We therefore demonstrate cross-modality generalization of \meow\ and its OW-learning capabilities using brain MRI data for tumor segmentation (using \brats\ \cite{antonelli22msd}). We summarize our results in \cref{tab:ow_brats}. \nninter\ and \sammed\ are excluded for this comparison since they were trained on brain MRI data.

\begin{table}[t]
\scriptsize
\centering
\caption{OW-IS results when learning within unknown MRI domain for brain tumors. All $\mathcal{F}$ were pre-trained on human CT data.}
\begin{tabular}{p{0.35cm}llll}
\hline
\textbf{ID} & \textbf{Method} & \textbf{m (\#SoIs)} & $\mathbf{DSC_{cr}^1\uparrow}$ & $\mathbf{DSC_{cr}^8\uparrow}$\\
\hline
C2.0 & \vista & \none & $0.291 \pm 0.0$ & $0.653 \pm 0.0$ \\
C2.1 & \meow\ ($\mathcal{F}_1 \& \mathcal{M}$) & $0$ & $0.555 \pm 0.006$ & $0.650 \pm 0.004$ \\
 &  & $\rightarrow 8$ & $0.635 \pm 0.013$ & $0.715 \pm 0.009$ \\
 &  & $\rightarrow 16$ & $0.642 \pm 0.014$ & $0.717 \pm 0.007$ \\
 &  & $\rightarrow 32$ & $0.640 \pm 0.001$ & $0.718 \pm 0.007$ \\
 &  & $\rightarrow 64$ & $0.643 \pm 0.003$ & $0.719 \pm 0.006$ \\
 &  & $\rightarrow 128$ & $0.648 \pm 0.010$ & $0.721 \pm 0.009$ \\
 C2.2 & \meow\ ($\mathcal{F}_2 \& \mathcal{M}$) & $0$ & $0.399 \pm 0.005$ & $0.551 \pm 0.005$ \\
 &  & $\rightarrow 8$ & $0.593 \pm 0.015$ & $0.616 \pm 0.020$ \\
 &  & $\rightarrow 16$ & $0.591 \pm 0.012$ & $0.612 \pm 0.022$ \\
 &  & $\rightarrow 32$ & $0.591 \pm 0.002$ & $0.606 \pm 0.016$ \\
 &  & $\rightarrow 64$ & $0.592 \pm 0.009$ & $0.613 \pm 0.013$ \\
 &  & $\rightarrow 128$ & $0.591 \pm 0.011$ & $0.613 \pm 0.013$ \\
 C2.3 & \meow\ ($\mathcal{F}_3 \& \mathcal{M}$) & $0$ & $0.674 \pm 0.001$ & $0.650 \pm 0.003$ \\
 &  & $\rightarrow 8$ & $0.725 \pm 0.022$ & $0.783 \pm 0.003$ \\
 &  & $\rightarrow 16$ & $0.726 \pm 0.019$ & $0.781 \pm 0.001$ \\
 &  & $\rightarrow 32$ & $0.715 \pm 0.011$ & $0.779 \pm 0.002$ \\
 &  & $\rightarrow 64$ & $0.723 \pm 0.001$ & $0.781 \pm 0.004$ \\
 &  & $\rightarrow 128$ & $\mathbf{0.729 \pm 0.007}$ & $\mathbf{0.787 \pm 0.001}$ \\
\end{tabular}
\label{tab:ow_brats}
\end{table}

Our approach outperforms \vista\ by a large margin when using $\mathcal{F}_1$ (C2.1) or $\mathcal{F}_3$ (C2.3), and when using $\mathcal{F}_2$ with $n<4$ clicks (C2.2). Although $\mathcal{F}_1$ has not been trained on either CT or MRI brain data, using it in combination with \meow\ still outperforms \vista, which has been trained on brain CT data. Highest performance is obtained when combining our method with \vista's sub-module ($\mathcal{F}_3$) while at the same time requiring less learnable parameters. For all feature extractors, we achieve much higher performance than the original \vista\ baseline already for $n=1$ clicks and $m=0$ memory samples. In agreement with our results for in-domain OW-learning, cross-domain OW-learning via memory consolidation further increases performance. Segmentation quality remains relatively stable with larger memory sizes, indicating a saturation effect (C2.3). In combination with qualitative demonstrations in \cref{fig:collage}, these results suggest that \meow\ generalizes across domains for OS-IS and OW-IS.

To summarize, we demonstrate that our approach flexibly adapts to various visual contexts without finetuning of learnable parameters. \meow\ succeeds in closed-set, open-set, and open-world settings while at the same time being modular, data-efficient, fast-learning, and feature-robust.

\section{Discussion}
\label{sec:discussion}

Our framework addresses the core challenges of OW-learning as formulated in \cref{method:openworld}. By leveraging memory consolidation within $\mathcal{M}$, the system facilitates seamless transition from zero-shot IS (via base prediction $y_b$) to few-shot and many-shot scenarios (via memory prediction $y_\mathcal{M}$) as knowledge is accumulated during deployment. Notably, our approach inherently supports incremental and continual learning paradigms, including out-of-distribution detection and data-efficient adaptation while keeping backbone parameters $\theta$ frozen. This enables scalable learning and unlearning without catastrophic forgetting \cite{mccloskey89forget,ratcliff90forget,french99forget,winwim24} by modulating memory states rather than performing computationally expensive gradient-based learning.

While performance in fully automated continual semantic- or panoptic segmentation frameworks tends to degrade over time due to error propagation \cite{kim24}, integrating a human user $\mathcal{U}$ in the loop provides a natural mechanism for pseudo-label quality verification. High-quality results may be learned while sub-optimal results may be discarded, ensuring purity of stored data for lifelong OW-learning. Through memory contents and different options for user guidance, our method further accounts for user-specific annotation styles, thereby enabling flexible personalization capabilities.

Moreover, the \meow\ architecture provides task-level flexibility. By reconfiguring the functional modules $\mathcal{F}$, $\mathcal{G}$, and $\mathcal{M}$, the system supports a broad range of downstream tasks, including instance classification, novelty detection, class labeling with arbitrary number of classes, semantic segmentation, IS, as well as HMP. Notably, this multi-task capability is achieved while maintaining a constant number of learnable parameters regardless of task complexity.

Recently proposed tasks for continual medical image segmentation include domain-, class-, and organ-level scenarios \cite{wang26contmed}. Notably, our method supports the domain scenario by consolidating images from different clinical centers while retaining the flexibility to adapt to specific domains by employing domain-specific memory modules. The class scenario is supported simply by swapping stored memory labels. The organ scenario is a combination of the other two, which is also supported due to our flexible architecture.

Deploying ML models in sensitive domains like medicine necessitates additional considerations regarding robustness, privacy, interpretability, and regulatory standards. Our parameter-free architecture offers several advantages in this regard:
\newline
\textbf{Regulatory compliance:} Our memory-focused architecture is compliant with mandates such as Europe's GDPR \textit{right to be forgotten} \cite[Article 17]{GDPR16}, which allows individuals to request that their personal data be removed under certain conditions. With our method, removal becomes as simple as deleting an entry from $\mathcal{M}$. This contrasts other monolithic methods, which often require complicated and inefficient unlearning or re-training schemes to remove the influence of specific data from neural network weights.
\newline
\textbf{Modularity and Maintenance:} The decoupled nature of $\mathcal{F}$, $\mathcal{G}$, and $\mathcal{M}$ allows flexible updates. Feature extractors can be swapped with better backbones or to address licensing changes while retaining the accumulated knowledge base. Memory modules can be swapped to specialize on specific classes or to meet user-specific preferences.
\newline
\textbf{Flexible Usage of Memory:} $\mathcal{M}$ may hold a fixed dataset with ground truth labels $z_{gt}$ of known classes to improve IS objectives. In the presence of $\mathcal{M}$, any interaction with $\mathcal{U}$ may be completely skipped to end up with a fully-automatic semantic- or instance segmentation system.
\newline
\textbf{Stability:} Our system seamlessly integrates any number of new classes. As $\mathcal{M}$ treats known classes independently for making predictions, the addition of new classes does not impact the segmentation performance on existing classes for KKCs and UKCs. Under the assumption of unconstrained memory, this translates to stable knowledge retention without catastrophic forgetting.
\newline
\textbf{Transparency and Usage:} By relying on parameter-free transduction with $\mathcal{G}$ and $\mathcal{M}$ for making predictions, our method is transparent and intuitive to use. Model predictions are governed by a set of interactive hyperparameters, each providing explicit control over a distinct conceptual aspect. This enables nuanced and predictable modifications to resulting segmentations.

\section{Conclusion}

We propose \meow, a modular open-world interactive segmentation (OW-IS) system combining a frozen feature extractor $\mathcal{F}$, a transduction module $\mathcal{G}$, and a memory module $\mathcal{M}$ to learn from its own predictions of unknown structures. Our approach enables flexible learning and unlearning paradigms through memory consolidation. The system takes weak user labels and interactive hyperparameters as guidance to segment any structure of interest. Predictions are made by leveraging hierarchical correspondence matching and message passing with label retrieval and transductive reasoning.

\textbf{Takeaways:} We demonstrate that non-interactive, non-OW backbones can be made interactive and OW-capable without any additional training of neural network parameters. By filling a memory module with self-labeled samples at test time, our system improves both in terms of segmentation performance and required user guidance. Notably, we show that our method is highly data-efficient and even works with completely untrained feature extractors. Due to our memory-based architecture, learning and unlearning is done in a matter of milliseconds, establishing a new standard for fast and stable learning at scale.

\textbf{Limitations:} Storing feature pyramids in $\mathcal{M}$ and applying $k$-NN search within memory contents generally requires more computational resources (higher hardware memory and slightly higher inference times) compared to fully end-to-end neural network based approaches. Moreover, because \meow\ requires feature extractors with specific hourglass-shaped architectures, it is currently unable to leverage transformer-based architectures \cite{dosovitskiy21vit} for open-world learning.

\textbf{Future Work:} Future directions include studying the effectiveness of self-supervised and unsupervised feature extractors. \meow\ can also be used to analyze sequential data such as longitudinal CT studies or videos. To enhance personalization, memory modules may be used to create user-specific profiles based on individual labeling preferences. Finally, to address privacy concerns inherent in handling sensitive data, privacy-preserving computation may be investigated.

Ultimately, this work provides a scalable foundation for deploying lifelong learning interactive segmentation systems capable of continuous self-improvement in dynamic real-world settings. 

\bibliographystyle{IEEEtran}
\bibliography{mainbib}

\newpage
\setcounter{section}{0}
\setcounter{page}{1}

\twocolumn[{
    \begin{center}
        \LARGE Appendix\\\meow: Medical Open-World Interactive Segmentation
    \end{center}
    \vspace{1.5cm}    
}]

\section{Method}
\label{app:method}

\begin{figure}[t]
	\centering
	\includegraphics[width=1\linewidth]{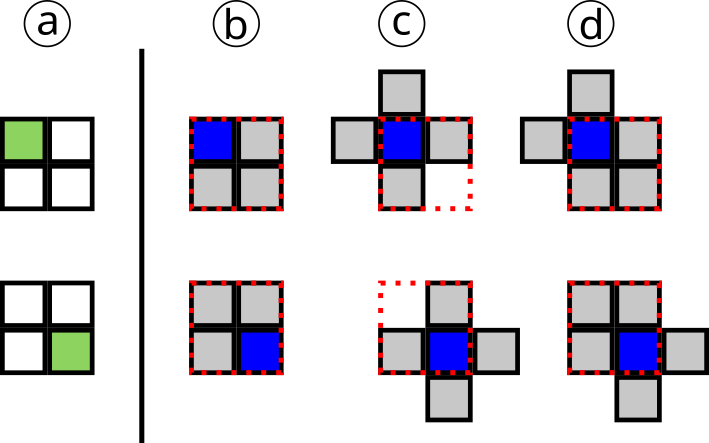}
	\caption{a) Query patch containing $2^d$ query nodes (in 2D for simplicity) that are used for similarity search. Green nodes indicate query nodes of interest. b) Original box-kernel with $2^d$ nodes, where a common kernel region is shared across all query nodes in the query patch. c) Newly introduced \textit{cross}-kernel with $1 + 2 \cdot d$ nodes, used for dense labels (e.g., message passing). Kernel positions (with anchor nodes in blue) adapt locally depending on relative query node positions in the patch. d) Newly introduced \textit{flexi}-kernel, used for sparse labels (e.g., weak user-labels). Kernel orientations adapt locally depending on query node positions. From the total of $2^d+d$ nodes within the kernel region, $2^d$ nodes are being kept while preferring labeled nodes as well as prioritizing nodes within the reference frame.}
	\label{fig:kernel}
\end{figure}

We visualize our newly introduced memory search kernels (called \textit{cross}-kernel and \textit{flexi}-kernel) in \cref{fig:kernel}, replacing the original box-kernel of \parm. In contrast to the former, where a common kernel region is shared across all query nodes that are related to a specific parent, locations and orientations of our new kernels adapt depending on query node positions. These adaptations can be regarded as location-informed biases, assuming that positional offsets in queries correlate to corresponding offsets in memory contents. The new kernels reduce metrication artifacts while requiring less (for \textit{cross}) or equal (for \textit{flexi}) hardware memory w.r.t. the box-kernel for 3D data (see ablation study results A0 vs. A2, A8 vs. A10, and A16 vs. A18 in the main text). Further, the \textit{flexi}-kernel improves search capacity by adapting to labeled regions in the image, effectively reducing overhead computations from unlabeled regions.

We provide pseudo-codes for calculating base predictions (\cref{alg:inference_step1}), fusing memory prediction (\cref{alg:inference_step2}), and hierarchical message passing (\cref{alg:inference_step3}).

\begin{algorithm}[t]
	\scriptsize
    \caption{Inference Step \#1: base prediction}
	\label{alg:inference_step1}
	\begin{algorithmic}
		\STATE $\text{Require }\text{hyperparameters: } (\Omega)^l, k_{start}, \phi$
        \STATE $\text{Require }\text{query pyramid: }(h)_q^l=\mathcal{F}(x_q)$
        \STATE $\text{Require }\text{user guidance: } u=\{z^\pm,\mathbf{\iota}\}\text{\hspace{0.1cm}};\text{\hspace{0.2cm}} \mathbf{\iota}=\{\beta_{reg}, \omega, \gamma,\xi^-, \beta_{nn}\}$
        \STATE $\text{Require }\text{working memory: } M=\{z_j, (h)_j^l\}$
		\STATE
        \STATE $\text{initialize}$:
		\STATE $m \gets \text{len}(M)$
        \STATE $k \gets k_{start}$
        \STATE $k_{use} \gets k_{start}$
        \STATE $id_{k,par} \gets p^n \times [1, 2, 3, ..., m]  \text{\hspace{2.75cm}parent index}$
		\STATE $s_{k,par} \gets p^n \times [1, 1, 1, ..., 1] \text{\hspace{3cm}parent similarity}$
        \STATE $(\mu)_j^l \gets \text{calc\_mask}(z^\pm, (\Omega)^l)  \text{\hspace{3.1cm}labeled regions}$
		\STATE
		\STATE $\text{hierarchical memory search}$:
		\FOR{\texttt{$l=L, 1$}}
            \STATE $id^l \gets \text{get\_children}(id_{k,par}, \mu_j^l, \text{kernel}) \text{\hspace{1.75cm}[$p^l \times k_{use} \cdot n_{ch}$]}$
            \STATE $o_q^l \gets \text{init\_meshgrid}(\Omega^l)$
            \STATE $s_{f,cos}^l \gets \text{cosine\_sim}(h_q^l, h_j^l[id^l])$ \text{\hspace{2.25cm}[$p^l\times k_{use} \cdot n_{ch}$]}
            \STATE $s_{f,match}^l \gets \text{softmax}_{p^l}(log(p^l)^{\beta_{reg}} \cdot s_{f,cos}^l) \text{\hspace{1cm}regularized matcher}$
            \STATE $s_{o,match}^l \gets 1-\text{L2dist}(o_q^l, o_j^l[id^l])/\text{L2dist\_max} \text{\hspace{0.75cm}[$p^l\times k_{use} \cdot n_{ch}$]}$
            \STATE $s_f^l \gets 0.5 \cdot (1 + s_{f,cos}^l)$
            \STATE $s_o^l \gets s_{o,match}^l$
            \STATE $s_{match}^l \gets \omega \cdot s_{f,match}^l + (1 - \omega) \cdot s_{o,match}^l$
            \STATE $s^l \gets \omega \cdot s_f^l + (1 - \omega) \cdot s_o^l \text{\hspace{2.cm}feature-proximity balancing}$
            \STATE $s_{match}^l[\neg \mu_j^l] \gets 0 \text{\hspace{3cm}search in labeled regions}$
            \STATE $s_{match}^l \gets s_{match}^l \cdot s_{k,par} \text{\hspace{1.95cm}update, [$p^l\times k_{use} \cdot n_{ch}$]}$
            \STATE $\Gamma \gets \gamma \cdot l / (n-1) + 0.5 \cdot (1 - \gamma)$
            \STATE $s^l \gets \Gamma \cdot s^l + (1 - \Gamma) \text{\hspace{2.6cm}local-global balancing}$
            \STATE $s^l \gets s^l \cdot s_{k,par} \text{\hspace{3.25cm}update, [$p^l\times k_{use} \cdot n_{ch}$]}$
            \STATE $id^l, s_{match}^l \gets \text{duplicate\_cleanup}(id^l, s_{match}^l)$		
            \IF{$l > 1$}
                \STATE $k \gets \lfloor\phi \cdot k\rfloor$
            \ENDIF
            \STATE $k_{use} \gets \text{min}(k, \sum\mu_j^l) \text{\hspace{3.05cm}upper bound}$
            \STATE $id_{keep} \gets \text{get\_topk}(s_{match}^l, k_{use}) \text{\hspace{1.75cm}[$p^l\times k_{use}$]}$
            \STATE $id_{k,par} \gets id^l[id_{keep}] \text{\hspace{3cm}update, [$p^{l-1}\times k_{use}$]}$
            \STATE $s_{k,par} \gets s^l[id_{keep}] \text{\hspace{3.25cm}update, [$p^{l-1}\times k_{use}$]}$
            \IF{$\text{init HMP graph}$}
                \STATE $i_j^l \gets id_{k,par}$
                \STATE $s_j^l \gets s_{k,par}$
                \IF{$l=1$}
                    \STATE $\text{return } (i)_j^l,\ (s)_j^l \text{\hspace{3.15cm}HMP graph}$
                \ENDIF
            \ENDIF
		\ENDFOR
		\STATE
		\STATE $\text{label-retrieval \& reduction}$:
		\STATE $id_{z} \gets \text{get\_children}(id_{k,par})$
		\STATE $y_j \gets z^\pm[id_z]$
        \STATE $s_j \gets s_{k,par}[id_z]$
        \STATE $w_j \gets p^1 \times [0.]  \text{\hspace{2.9cm}init weights}$
        \STATE $id^+ \gets y_j = +1 \text{\hspace{2.8cm}positives}$
        \STATE $id^- \gets y_j = -1 \text{\hspace{2.8cm}negatives}$
        \STATE $w_j[id^+] \gets 1$
        \STATE $w_j[id^-] \gets \xi^- \text{\hspace{3cm}negative click strength}$
        \STATE $w_j \gets \text{softmax}_k(\beta_{nn} \cdot w_j \cdot s_j) \text{\hspace{1.3cm}commitment}$
        \STATE $y_b^\pm \gets \sum_{j=1}^k w_j \cdot s_j \cdot y_j$
        \STATE $y_b \gets 0.5 \cdot (1 + y_b^\pm)$
        \STATE
        \STATE $\text{return } y_b \text{\hspace{1cm}base prediction} \in \{ y \mid 0 \leq y \leq 1 \}$
	\end{algorithmic}
\end{algorithm}

\begin{algorithm}[t]
	\scriptsize
        \caption{Inference Step \#2: fusing memory prediction}
	\label{alg:inference_step2}
	\begin{algorithmic}
		\STATE $\text{Require } \tau, y_b, \alpha_{fuse}, \text{optional: }y_M$
		\STATE
            \IF{$\exists\ y_M$}
                \STATE $\tau^\pm \gets 2 \cdot \tau - 1$ 
                \STATE $y_b^\pm \gets 2 \cdot y_b - 1$
                \STATE $y_M^\pm \gets 2 \cdot y_M - 1$
                \STATE $y_M^\pm \gets \text{CDF\_matching}(y_M^\pm, y_b^\pm, \alpha_{fuse})$
                \STATE $id^+ \gets y_b^\pm > \tau^\pm \text{\hspace{3.45cm}mask of positives}$
                \STATE $id^- \gets y_b^\pm < -\tau^\pm \text{\hspace{3.25cm}mask of negatives}$
                \STATE $y_r^\pm \gets y_M^\pm \text{\hspace{4.3cm}init (incl. neutral)}$
                \STATE $y_r^\pm[id^+] \gets \text{max}(y_b^\pm[id^+], y_M^\pm[id^+])$
                \STATE $y_r^\pm[id^-] \gets \text{min}(y_b^\pm[id^-], y_M^\pm[id^-])$
                \STATE $y_r \gets 0.5 \cdot (1 + y_r^\pm)$
            \ELSE
                \STATE $y_r \gets y_b$
            \ENDIF
            
            \STATE
            \STATE $\text{return } y_r \text{\hspace{1cm}raw prediction} \in \{ y \mid 0 \leq y \leq 1 \}$
	\end{algorithmic}
\end{algorithm}

\begin{algorithm}[t]
	\scriptsize
        \caption{Inference Step \#3: hierarchical message passing}
	\label{alg:inference_step3}
	\begin{algorithmic}
		\STATE $\text{Require } y_r, \tau, \sigma_{mp}, n_{mp}, \eta_{mp}, \lambda_{mp}, \beta_{mp}, (\Omega)^l, (s_j)^l, (i_j)^l$
		\STATE
            \STATE $\nu \gets \text{voxel\_count}(y_r, \tau)  \text{\hspace{3.65cm}segment size}$
            \STATE $y \gets \text{clamp}(\sigma_{mp} \cdot (y_r - \tau) + \tau, \text{min}=0, \text{max}=1) \text{\hspace{0.5cm}sharpen}$
            \STATE
            \FOR{$l=L_{mp},1$}
                \STATE $y \gets \text{interpolate}(y, \Omega^l)$
                \STATE $z_l \gets \text{interpolate}(z^\pm, \Omega^l)$
                \STATE $id^+ \gets z_l > 0.5$
                \STATE $id^- \gets z_l < -0.5$
                \STATE $e_j^l \gets \text{softmax}_\kappa(\beta_{mp} \cdot s_j^l) \text{\hspace{3.1cm}graph edges}$
                \FOR{$\text{step}=1,\eta_{mp}$}
                    \STATE $y[id^+] \gets 1$
                    \STATE $y[id^-] \gets 0$
                    \STATE $y_{msg} \gets \sum_\kappa e_j^l \cdot y[i_j^l]  \text{\hspace{3cm}aggregated message}$
                    \STATE $y \gets \lambda_{mp} \cdot y_{msg} + (1 - \lambda_{mp}) \cdot y \text{\hspace{1.7cm}update}$
                \ENDFOR
            \ENDFOR
            \STATE
            \STATE $\tau \gets \text{sort}(y)[\nu]  \text{\hspace{4.7cm}conserve size}$
            \STATE
            \STATE $\text{return } y \text{\hspace{1cm}final prediction} \in \{ y \mid 0 \leq y \leq 1 \}$
	\end{algorithmic}
\end{algorithm}

\section{Data}
\label{app:data}

\subsection{Datasets}
Following, we briefly describe all datasets that are used in this work:
\newline
\textbf{\msd \cite{antonelli22msd}:} The Medical Segmentation Decathlon combines 10 different medical segmentation datasets. We use whole-organ CT annotations for liver, pancreas, and spleen as KKCs. When using \totseg\ for feature extraction, we use \msd\ as an independent evaluation set.
\newline
\textbf{\kits \cite{heller23kits}:} We use whole-organ annotations for left and right kidneys from the Kidney Tumor Segmentation challenge as KKCs. When using \totseg\ for feature extraction, we use \kits\ as an independent evaluation set.
\newline
\textbf{\mousemicro \cite{rosenhain18}:} We use organ annotations from 140 native and 85 contrast-enhanced, whole-body mouse $\mu$CT scans. Notably, the acquired resolution significantly exceeds that of typical human CT data, achieving a  voxel spacing of $0.14\ mm$. W.r.t. human CT data, the \mousemicro\ exhibit a strong domain shift both in terms of spatial scale and content. In our OW experiments, mouse organ classes are considered UKCs both for our method and baseline methods. Classes for the native set include heart and lung, while classes for the contrast-enhanced set include spleen, lung, heart, kidneys, bladder, stomach, liver, and intestine.
\newline
\textbf{\antbrain \cite{toulkeridou23insect}:} We use ant brain annotations from 143 (118 training, 25 testing) $\mu$CT scans from different species. Again, the acquired resolution significantly exceeds that of typical human CT data, achieving isotropic voxel spacings between $1-10\mu m$. In our pipeline, data is min-max normalized and resampled to $7.5\mu m$ for all methods.
\newline
\textbf{\totsegdata \cite{wasserthal23totseg}:} We compile a subset of \totseg\ data comprising liver, pancreas, spleen, kidney left, and kidney right for ablation studies. When using \suprem\ for feature extraction, we use \totsegdata\ as an independent evaluation set.
\newline
\textbf{\brats \cite{antonelli22msd}:} We use \brats\ MR brain segmentation data to investigate cross-domain generalization for OW-IS.

\subsection{Data Preparation}
We distinguish between 3 data split types: train hyper (for hyperparameter optimization), train memory (for model training via memory consolidation), and validation. For \msd+\kits, we employ an $80:20$ split and partition the training split into a $10\%$ train hyper and a $70\%$ train memory sub-split. \totsegdata\ and \antbrain\ datasets have official training and validation splits. We partition the \totsegdata\ training split into a $6.25\%$ train hyper and a $93.75\%$ train memory sub-split. For \mousemicro\ and \brats, we apply a $70:30$ training vs. validation split. Note that \mousemicro, \antbrain, and \brats\ do not have train hyper splits because we treat them as unknown in our open-set and open-world experiments. Thus, we do not perform any hyperparameter optimization on them.

We apply a folding strategy for all open-world experiments to study the impact of stored memory samples on performance. For \mousemicro, \antbrain, and \brats, we partition the training split into 2 independent, non-overlapping folds. We do not apply folding of the validation split. This setup allows for memory sizes of $m\le128$ SoIs for any training fold.

Medical data (\msd, \kits, \totsegdata, and \brats) is canonicalized and resampled to a spacing of $1.5\ mm$. We use a crop size of $\Omega^1=(128\times128\times128)$ voxels for all experiments. Window-levels for \suprem\ and \vista\ are $[-175, 250]$ and $[-60, 140]$, respectively. For \totseg\ (organs model), we follow the official pre-processing scheme according to available meta-information \cite{wasserthal23totseg}. Data is z-normalized for \sammed\ and not normalized for \nninter\ in analogy to their official documentation.

\section{Virtual User}
\label{app:virtual_user}
For automated evaluation, we employ a virtual user $\mathcal{U}_v$ that simulates the typical behavior of a human user $\mathcal{U}_h$. For a given dataset and IS setup, $\mathcal{U}_v$ simulates clicks under fixed hyperparameter settings. The initial click is always placed at the volume center where the SoI resides. Afterwards, $\mathcal{U}_v$ executes the correction click strategy described in Algorithm \ref{algo:click_policy_virtual_user}, where clicks are placed within a RoI depending on over- or under-segmentation. Given that $\mathcal{U}_h$ rarely clicks at the edge of a ground-truth mask $z_{gt}$ unless necessary, $\mathcal{U}_v$ similarly prefers placing clicks far away from the edge. For efficient selection of click positions, we sample 32 random candidates from the RoI and select the one farthest away from the RoI's edge. Given a single-pixel click, we extend it to a 7-voxel cross-shaped click region for \meow.

\begin{algorithm}[H]
\scriptsize
\caption{click-policy of virtual user $\mathcal{U}_v$}\label{algo:click_policy_virtual_user}
\begin{algorithmic}
\STATE $\text{inputs: }y, z_{gt}, z^\pm, \tau$
\STATE $FN \gets y < \tau \cup z_{gt} = 1 \cup z^\pm \ne 1 \text{\hspace{1.5cm}false negatives}$
\STATE $FP \gets y \ge \tau \cup z_{gt} = 0 \cup z^\pm \ne -1 \text{\hspace{1.33cm}false positives}$
\IF {$\sum{FP} > \sum{FN}$}
    \STATE $\text{click} \gets \text{"negative"}$
    \STATE $RoI \gets FP$
\ELSE
    \STATE $\text{click} \gets \text{"positive"}$
    \STATE $RoI \gets FN$
\ENDIF
\STATE $p_j \gets \text{random\_position}(RoI) \text{ for } j = 1,...,32$
\STATE $d_j \gets \text{calc\_distance}(p_j, \text{ edge}(RoI))$
\STATE $i = \text{argmax}(d_j)$
\STATE \text{return} $\text{click},\  p_j[i]$
\end{algorithmic}
\end{algorithm}

\section{Hyperparameters}
\label{app:hyperparam_optim}

For adapting to specific use cases and diverse annotation preferences of different users, we introduce several components to increase overall flexibility of our method. These components can be controlled independently and dynamically (e.g., via sliders) as interactive hyperparameters $\boldsymbol{\iota}$, allowing $\mathcal{U}$ fine-grained control over segmentation behavior (also see Algorithms 1-3):
\begin{itemize}
    \item Classification Threshold: A threshold $\tau \in [0, 1]$ is set to discriminate between foreground and background. High values are more conservative and lead to smaller segments.
    \item Commitment -- Exploration: $\beta_{reg} \in [-1, 0]$ influences the extent of exploration during correspondence matching by making stored patterns more exclusive. A $\beta_{reg}$ close to zero means that stored patterns are more exclusive, in turn leading to more exploration.
    \item Similarity Score: The significance of spatial proximity compared to feature-similarity (both w.r.t. $z^\pm$) is adjusted with $\omega \in [0, 1]$. High values prioritize feature-similarity over proximity.
    \item Balancing: The balancing parameter $\gamma\in (-1, 1)$ controls how much emphasis is put on local, high-res features versus global, low-res features. Low values prioritize local features while high values prioritize global features.
    \item Negative Labels: $\xi^- \in [0, 1]$ determines the overall strength of negative labels obtained by negative clicks or brushes, with $\xi^-=1$ having equal contribution of $z^+$ and $z^-$. 
    \item Commitment -- Sparse Labels: $\beta_{nn} \in [0.1, 1000]$ influences the extent of more similar NNs contributing to base predictions $y_b$. While large $\beta_{nn}$ focuses on the most similar NN, small $\beta_{nn}$ assigns equal importance to all NNs.
    \item Fusion Strategy: $\alpha_{fuse} \in [0, 1]$ determines how much to blend from the source distribution $y_M^\pm$ towards the reference distribution $y_b^\pm$ before fusing them into $y_r$ (see \cref{alg:inference_step2}).
    \item HMP Sharpening: $\sigma_{mp}\in [1, 25]$ determines sharpening strength for mitigating over-smoothing during HMP.
    \item Commitment -- HMP: $\beta_{mp} \in [0.1, 10]$ influences the extent of more similar NNs contributing to messages. While large $\beta_{mp}$ increase graph sparsity, small $\beta_{mp}$ distribute equal weight to all NNs.
    \item HMP Strength: $\lambda_{mp}\in [0, 1]$ is the update strength for all HMP levels, with $\lambda_{mp}=0$ fully disabling HMP.
\end{itemize}

\textbf{Hyperparameter Optimization:} To find an optimal set of interactive and non-interactive hyperparameters, we leverage Bayesian optimization \cite{shahriari16bayes}, a probabilistic approach to maximize an objective function $f(\textbf{x})$ over a bounded parameter space $\mathcal{X}$. In our case, $f(\textbf{x})$ maps a set of hyperparameters $\textbf{x}\in\mathcal{X}$ to the resulting IS performance over an evaluation split. 
Optimization is performed using Scikit-Optimize \cite{scikit-optimize20} (version 0.10.2) with default settings. We use 10 initial, random evaluation steps, followed by optimization until convergence. The Mat\'ern kernel \cite{matern60} is chosen to model the smoothness of $f(\textbf{x})$. We optimize the following parameters within respective intervals: $\tau \in [0.7, 0.95],\ \omega \in [0.5, 0.9],\ \gamma \in [-0.99, 0],\ \xi^- \in [0.025, 0.5],\ log(\beta_{nn}) \in [-1., 3.],\ \beta_{reg} \in [-1, 0],\ \alpha_{fuse} \in [0.75, 1],\ \sigma_{mp} \in [1, 25],\ \lambda_{mp} \in [0, 1],\ L_{mp} \in [1, 3],\ log(\beta_{mp}) \in [-1., 1.],\ \eta_{mp} \in [6, 12],\ \text{TTA} \in [\xmarkraw, \cmarkraw]$. Optimal hyperparameters for all feature extractors $\mathcal{F}$ are given in \cref{tab:hyperparams}.

\begin{table}[t]
\scriptsize
\centering
\caption{Results for Bayesian hyperparameter optimization using KKC organ classes liver, pancreas, spleen, and kidney.}
\setlength{\tabcolsep}{3.5pt}

\rotatebox{90}{
\begin{tabular}{lllcccccccccccccc}
\hline 
\textbf{ID} & \textbf{Data} & \textbf{$\mathcal{F}$} & \textbf{Mem} & $\tau$ & $\omega$ & $\gamma$ & $\xi^-$ & $log(\beta_{nn})$ & $\beta_{reg}$ & $\alpha_{fuse}$ & $\sigma_{mp}$ & $\lambda_{mp}$ & $L_{mp}$ & $log(\beta_{mp})$ & $\eta_{mp}$ & TTA \\
\hline
H0 & \totsegdata & $\mathcal{F}_1$ & \cmarkraw& 0.873& 0.900& -0.850& 0.120& -1.000& 0.000& 0.000& 1.000& 0.957& 2& 0.462& 12& \xmarkraw\\
H1 & \totsegdata & $\mathcal{F}_1$ & \xmarkraw& 0.821& 0.900& -0.990& 0.500& 3.000& \none& \none& 22.583& 1.000& 2& -1.000& 12& \xmarkraw\\
H2 & \msd+\kits & $\mathcal{F}_1$ & \cmarkraw& 0.863& 0.900& -0.990& 0.462& 1.308& 0.000& 0.000& 25.000& 1.000& 2& 0.805& 12& \xmarkraw\\
H3 & \msd+\kits & $\mathcal{F}_1$ & \xmarkraw& 0.812& 0.900& -0.990& 0.025& 2.516& \none& \none& 21.581& 1.000& 3& 1.000& 12& \xmarkraw\\
H4 & \totsegdata & $\mathcal{F}_2$ & \cmarkraw& 0.842& 0.900& -0.179& 0.177& 0.113& -1.000& 0.000& 1.000& 0.718& 2& 1.000& 6& \xmarkraw\\
H5 & \totsegdata & $\mathcal{F}_2$ & \xmarkraw& 0.818& 0.900& -0.990& 0.089& 1.516& \none& \none& 25.000& 0.441& 3& 0.310& 6& \xmarkraw\\
H6 & \msd+\kits & $\mathcal{F}_2$ & \cmarkraw& 0.860& 0.500& -0.990& 0.025& -0.851& -0.709& 0.000& 25.000& 1.000& 2& 1.000& 6& \xmarkraw\\
H7 & \msd+\kits & $\mathcal{F}_2$ & \xmarkraw& 0.813& 0.900& -0.990& 0.500& 3.000& \none& \none& 25.000& 0.732& 3& 1.000& 6& \xmarkraw\\
H8 & \totsegdata & $\mathcal{F}_3$ & \cmarkraw& 0.827& 0.500& -0.990& 0.025& -0.524& -0.349& 0.000& 6.218& 0.427& 2& -0.381& 12& \cmarkraw\\
H9 & \totsegdata & $\mathcal{F}_3$ & \xmarkraw& 0.786& 0.682& -0.990& 0.435& 2.395& \none& \none& 25.000& 0.799& 3& 0.125& 6& \cmarkraw\\
H10 & \msd+\kits & $\mathcal{F}_3$ & \cmarkraw& 0.841& 0.651& -0.549& 0.025& 2.587& -0.434& 0.022& 1.000& 0.723& 2& 0.275& 12& \cmarkraw\\
H11 & \msd+\kits & $\mathcal{F}_3$ & \xmarkraw& 0.775& 0.707& -0.990& 0.497& 3.000& \none& \none& 25.000& 0.800& 3& -0.197& 10& \cmarkraw\\
\end{tabular}
}
\label{tab:hyperparams}
\end{table}

\clearpage

\section{List of Symbols and Abbreviations}
\label{app:list_of_symbols}

{\small
\begin{tabular}{ll}
  \textbf{Notation} & \textbf{Description} \\
  \hline
  $\alpha_{fuse}$ & strength of CDF matching\\
  $\beta_{mp}$ & Hopfield commitment (message passing)\\
  $\beta_{nn}$ & Hopfield commitment ($k$-NN search)\\
  $\beta_{reg}$ & Hopfield commitment (matcher)\\
  $c$ & class index \\
  CT & Computed Tomography\\
  $DSC$ & Dice-Sørensen coefficient\\
  $(e)^l$ & message passing edge weights\\
  $\eta_{mp}$ & number of message passing steps\\
  $\mathcal{F}$ & feature extractor (neural network)\\
  $\gamma$ & balancing parameter (global vs. local)\\
  $\mathcal{G}$ & transduction module\\
  HMP & hierarchical message passing\\
  $(h)^l$ & multi-scale feature pyramid\\
  $id$ & voxel indices (pseudo-code)\\
  $(i)^l$ & voxel indices (message passing connectivity)\\
  IC & instance classification\\
  IS & interactive segmentation\\
  $\mathbf{\iota}$ & interactive hyperparameters\\
  $k$ & number of nearest neighbor voxels\\
  $\kappa$ & number of nearest neighbor HMP nodes\\
  $\mathsf{k}$ & number of nearest neighbor patches\\
  KKC & known known class\\
  KUC & known unknown class\\
  $L$ & number of feature pyramid levels\\
  $L_{mp}$ & number of message passing levels\\
  $L_{sp}$ & number of sparse levels\\
  $\lambda_{mp}$ & message passing strength\\
  $m$ & number of memory samples\\
  $M$ & working memory\\
  $\mathcal{M}$ & memory module\\
  MR & magnetic resonance\\
  $(\mu)^l$ & labeled regions of feature pyramid\\
  $n$ & number of clicks \\
  ND & novelty detection \\
  $NN$ & nearest neighbor\\
  $NoC$ & number of clicks metric\\
  $\Omega$ & spatial domain (spatial resolution)\\
  $\omega$ & balancing parameter (feature vs. position)\\
  OS & open set\\
  OW & open world\\
  $p$ & number of voxels \\
  RoI & region of interest\\
  $\mathcal{S}$ & interactive segmentation model\\
  $\sigma_{mp}$ & message passing sharpening\\
  SoI & segment of interest\\
  $\tau$ & classification threshold\\
  $\theta$ & learnable parameters of $\mathcal{F}$\\
  $t$ & time\\
  TTA & test-time augmentation\\
  $u$ & user guidance \\
  $\mathcal{U}$ & user\\
  UKC & unknown known class\\
  UUC & unknown unknown class\\
  $w$ & weight\\
  $x$ & model input\\
  $\xi^-$ & negative click strength\\
  $y$ & prediction\\
  $z^\pm$ & weak user labels (mask)\\
  $z_{gt}$ & ground truth labels (mask)\\
  $z_p$ & pseudo-labels (mask)\\
  $\zeta$ & novelty threshold \\
\end{tabular}
}

\end{document}